\documentclass[sigconf,table]{acmart}
\renewcommand\footnotetextcopyrightpermission[1]{} % removes footnote with conference information in first column
\usepackage{graphicx}
\usepackage{adjustbox}
\usepackage{amsmath}
\usepackage{centernot}
\usepackage{multirow}
\usepackage{booktabs}
\usepackage{subfigure}
\usepackage{caption}
\usepackage{dsfont}
\usepackage{enumerate}
\usepackage{enumitem}
\usepackage{empheq}
\usepackage{amsfonts}
\usepackage{totcount}
\usepackage{makecell}
\usepackage{xcolor}
\usepackage{changepage}
\usepackage{stfloats}
\usepackage{soul}
\usepackage[ruled,vlined]{algorithm2e}
\usepackage{algpseudocode}
\usepackage[title]{appendix}
\usepackage{pdfpages}

\begin{document}

\title{Realization of Causal Representation Learning to Adjust Confounding Bias in Latent Space}

%\author{Anonymous}
% \affiliation{%
\fancyhead{}
\author{Jia Li$^{1}$,  Xiang Li$^2$, Xiaowei Jia$^3$, Michael Steinbach$^1$, Vipin Kumar$^1$\\
\small\baselineskip=9pt University of Minnesota, $^1$ Computer Science;
\small\baselineskip=9pt $^2$ Bioproducts and Biosystems Engineering. 
\small\baselineskip=9pt $^3$ University of Pittsburgh, Computer Science.\\
\small  $^1$ \{jiaxx213, stei0062, kumar001\}@umn.edu,$^2$ lixx5000@umn.edu,
\small $^3$ xiaowei@pitt.edu
}

\begin{abstract}

Causal DAGs(Directed Acyclic Graphs) are usually considered in a 2D plane. Edges indicate causal effects' directions and imply their corresponding time-passings. Due to the natural restriction of statistical models, effect estimation is usually approximated by averaging the individuals' correlations, i.e., observational changes over a specific time.
However, in the context of Machine Learning on large-scale questions with complex DAGs, such slight biases can snowball to distort global models - More importantly, it has practically impeded the development of AI, for instance, the weak generalizability of causal models.
In this paper, we redefine causal DAG as \emph{do-DAG}, in which variables' values are no longer time-stamp-dependent, and timelines can be seen as axes. By geometric explanation of multi-dimensional do-DAG, we identify the \emph{Causal Representation Bias} and its necessary factors, differentiated from common confounding biases.
Accordingly, a DL(Deep Learning)-based framework will be proposed as the general solution, along with a realization method and experiments to verify its feasibility.

% Involved causal data includes Electronic Healthcare Records (EHR) to estimate medical effects and a hydrology dataset to forecast the environmentally influenced streamflow.

\end{abstract}

\keywords{Causal Representation, Causal Graph, DAG, do-DAG, do-Calculus, Geometric Explanation of do-DAG, Higher-Dimensional Representation, Multi-Timelines in Causality, Deep Learning.}

\maketitle

%\vspace{-3mm}

\section{Introduction}
\label{sec:intro}
Causality is a fundamental study in many fields, like meteorology, biology, epidemiology, economics, social sciences, etc. \cite{wood2015lesson,vukovic2022causal,ombadi2020evaluation}.
It aims to uncover the underlying causal relationship, which generated the observational data.%, to answer questions like ``when the flood will happen'', ``whether the medication works well'', and so on.
The study of causal inference originated from classical statistics and has helped to construct considerable domain knowledge in past decades.
But in recent years, technical progresses dramatically sped up data collecting and brought it a significant challenge: learning causal structure from data as Bayesian networks is NP-hard. On the other hand, Machine Learning (ML) from computer science shows advanced effectiveness in handling big data, and its developments on causality\cite{scheines1997introduction, ahmad2018interpretable, sanchez2022causal} have been considered to constitute the bedrock of Artificial Intelligence (AI) \cite{li2020causal}.

Causal DAG (Directed Acyclic Graph)\cite{pearl2009causal} in causal inference is the concept carrying causal knowledge. It describes probabilistic dependencies among variables to represent causations.
Thus, from a modeling view, causality is equal to correlations with a specified direction; but from a knowledge-driven view, it supposes to be much richer. Considerable works in various fields detected their connection further to describe causality as multi-variates correlations or correlated time series
\cite{guyon2008practical, zhao2013gut, marwala2015causality}. But essentially, they are improving the approximation of causality by using correlations.

Since ML provides more global views via overall optimizations, such a conceptual gap becomes more evident, especially when the knowledge DAG has larger scales or higher complexity.
Typically, in recent Deep Learning (DL) applications, the ``DAG-ness'' constraint has been quantified as continuous functions\cite{zheng2018dags, zheng2020causal, lachapelle2019gradient}, realized the global structural optimization without NP-hard concern; but the attempts of shaping architectures according to DAGs cannot promisingly benefit the modeling accuracy \cite{ma2018using, zheng2020causal, luo2020causal}, which implies the correlation-built AI fail to fit with the ground truth causality.

% it feels strange that they are still far from practical use in many domains, like healthcare or bioinformatics \cite{qayyum2020secure, chen2021ethical, lecca2021machine}.
% A growing consensus is that ML models are more powerful, but statistics can keep models interpretable well \cite{crown2019real}.
% However, understanding why model effectiveness and interpretability can be paradoxical is not intuitive.

Furthermore, the generalization problem of causal models has drawn attention\cite{scholkopf2021toward}. It implies our way of modeling causal variables likely biases the accurate representation of causal knowledge, which has not bothered causal inference so profoundly. Nevertheless, without solving it, the causal models' individualization can also be on hold due to the risks of becoming overfitting
\cite{salman2019overfitting} - Even though we know both of them are much needed in domain sciences, like healthcare or bioinformatics\cite{qayyum2020secure, chen2021ethical, lecca2021machine}.

We ever proved the existence of the inherent bias in causality DL\cite{li2020teaching}. Accordingly, this paper aims to identify such bias, often mixed with common confounding biases, named as \emph{Causal Representation Bias} (CRB).
The source of CRB traces back to the definition of causal DAG and is attributable to the conceptual confusion of correlation and causality.
For example, taking medicine $M$ expects $30$ days to achieve its full effect; but a young patient $P_1$ only used $20$ days, and another older patient $P_2$ used $50$ days. The 30-day status changes collected from them are individual-level \emph{correlations}; but as \emph{effects}, they are $150\%$ and $60\%$, respectively. 
Averaging correlations to approximate the effect is not a problem in the single-effect estimation (like the typical scenario of statistical causal inference), 
but when multiple effects form a complex graph, such bias will be produced in each effect estimation and snowball to distort the global models (like the typical scenario in causality DL). 

Like what Pearl emphasized in do-calculus\cite{pearl2012calculus}, observation and intervention are two things.
Modeling on the former is to obtain correlations, while the latter is for causal effects.
Based on this, we redefine causal DAG as \emph{do-DAG}, where the nodes' values are no longer time-stamp-dependent. Then, timelines can be considered independent axes, bringing new geometric meanings, which leads to conclude the three necessary factors of CRB: \emph{confounding}, \emph{multi-timelines}, and \emph{data heterogeneity}.
Realizing do-DAG relies on DL-based representation extraction. Variables can be reconstructed from latent space vectors, representing the individuals' status instead of observed values. Accordingly, the new framework \emph{Causal Representation Learning} (CRL) will be introduced, as well as its realization method and the experiments to verify the feasibility.

As principal contributions of this paper, we:
\begin{itemize}
    \item Raise the do-DAG concept and its geometric explanation;
    \item Identify the source of CRB and propose CRL;
    \item Propose a novel architecture to extract higher-dimensional representation for realizing CRL.
\end{itemize}

\subsection{do-DAG: the Redefined Causal DAG}
In this section, we will redefine the concept of causal DAG (Directed Acyclic Graph) and discuss its geometric meaning accordingly.

In statistical causal inference\cite{pearl2009causal}, pair-wise causation $A \rightarrow B$ means: 1) variables $A$ and $B$ are associated, and 2) this association has specific time order - ``A changes'' must before ``B changes''. The role of $B$ is inherently ambiguous - Is it a variable? Or $A$'s effect? 

According to SEMs (Structural Equation Models), variable $B$ is defined as $B=f(A)$, where function $f$ describes the correlation between $A$'s value at time $t_A$ and $B$'s value at time $t_B$, given that $t_B>t_A$ are two time-stamps with a pre-determined timespan between them.
However, $f(A)$ may not define the causal effect ``what is $B$'s (must-happening) value changing after $A$ changes?''
For example, in the above case, correlation = [status changes of $P_1$ and $P_2$ after $30$ days], where $P_1$ has achieved $100\%$ expected medical effects of $M$, but $P_2$ only achieved $60\%$; while causal effect = [the expected ($100\%$) status changing after taking $M$ for any patient].

SEMs assume the observational correlation fully represents desired causal effects, which brings another trouble - the hidden confounder. In this case, the $40\%$ difference between $P_1$ and $P_2$ is ascribed to the unobservable variable, which identifies the individual-level body features (including the aging factor), differing among patients $P_1$ and $P_2$. To distinguish two concepts, Pearl raises \emph{do-calculus} \cite{pearl2012calculus} in 2012, where $A$ and $do(A)$ indicate: the observational values of variable $A$, and, the intervention action ``$A$'s value changing happens'', respectively. Correspondingly, $B$'s two roles can be differentiated: $B=f(A)$ is a correlated variable valued by time stamps, and $B=f(do(A))$ is the causal effects, whose value is determined by action $do(A)$ only and regardless of time stamps.

Significantly, we redefine $A \rightarrow B$ to denote the effects $B=f(do(A))$ only, such that any directed edge in causal DAGs is no longer a function of time-stamp $t$, named \underline{\emph{do-DAG}} to identify.

In do-DAG, $G=(\mathcal{V},\mathcal{E})$, vertices $\mathcal{V}$ (i.e., nodes) are variables representing individual status, whose values changing reflect causal effects, determined by the structure of edges $\mathcal{E}$. 
E.g., Let $V \in \mathcal{V}$ have two parents $A, B \in \mathcal{V}$, whose value is an $m$-dimentional vector, $v \in \mathbb{R}^m$, then $v=v_1+v_2=f_1(a)+f_2(b), v_1, v_2 \in \mathbb{R}^m$; Here $a$ and $b$ are the values of $A$ and $B$ corresponding to $v$, and $f_1, f_2$ represent the effects delivered by edges $e_1=\overrightarrow{AV}$ and $e_2=\overrightarrow{BV}$, where $e_1, e_2 \in \mathcal{E}$.
%Notably, we use $f(\cdot)$ to denote $f(do(\cdot))$ for briefty.

The first advantage of do-DAG is that it can significantly simplify causal graphs by avoiding additional hidden confounders; Also, the do-calculus can become more intuitively explainable. 
The classical models from statistics are inherently restricted and cannot implement variables as status-representing
detached from time stamps. But nowadays, DL-based autoencoders can realize such variables as representation vectors in latent space, and accordingly, the individual-level features present to be different time-passing speeds for different individuals in the data population. Notably, in do-DAG, ``timeline'' has become an independent concept necessarily used to identify the causal effect on edge, so we need to establish the geometric explanation of do-DAG from the beginning.

%\vspace{-5mm}
\subsection{Geometric Explanation of do-DAG}

%\vspace{2mm}
\noindent
\emph{\textbf{Definition 1.} A \underline{timeline} consists of successive time stamps split uniformly, along which a causal effect function $f(\cdot)$ is defined.}

% \vspace{2mm}
% \noindent
% \emph{\textbf{Definition 2.} A \underline{learning plane} consists of two axes: 1) learning timeline $T$, and 2) the axis of different nodes $\mathcal{V}$ nominally separated.}

%\vspace{2mm}
\noindent
\emph{\textbf{Definition 2} A \underline{do-DAG space} is defined by all axes of relatively independent timelines, and also the axis where different nodes $\mathcal{V}$ are nominally separated.}
\vspace{2mm}

\begin{figure}[hbp]
\vspace{-2mm}
\hspace*{-7mm}
\includegraphics[scale=0.47]{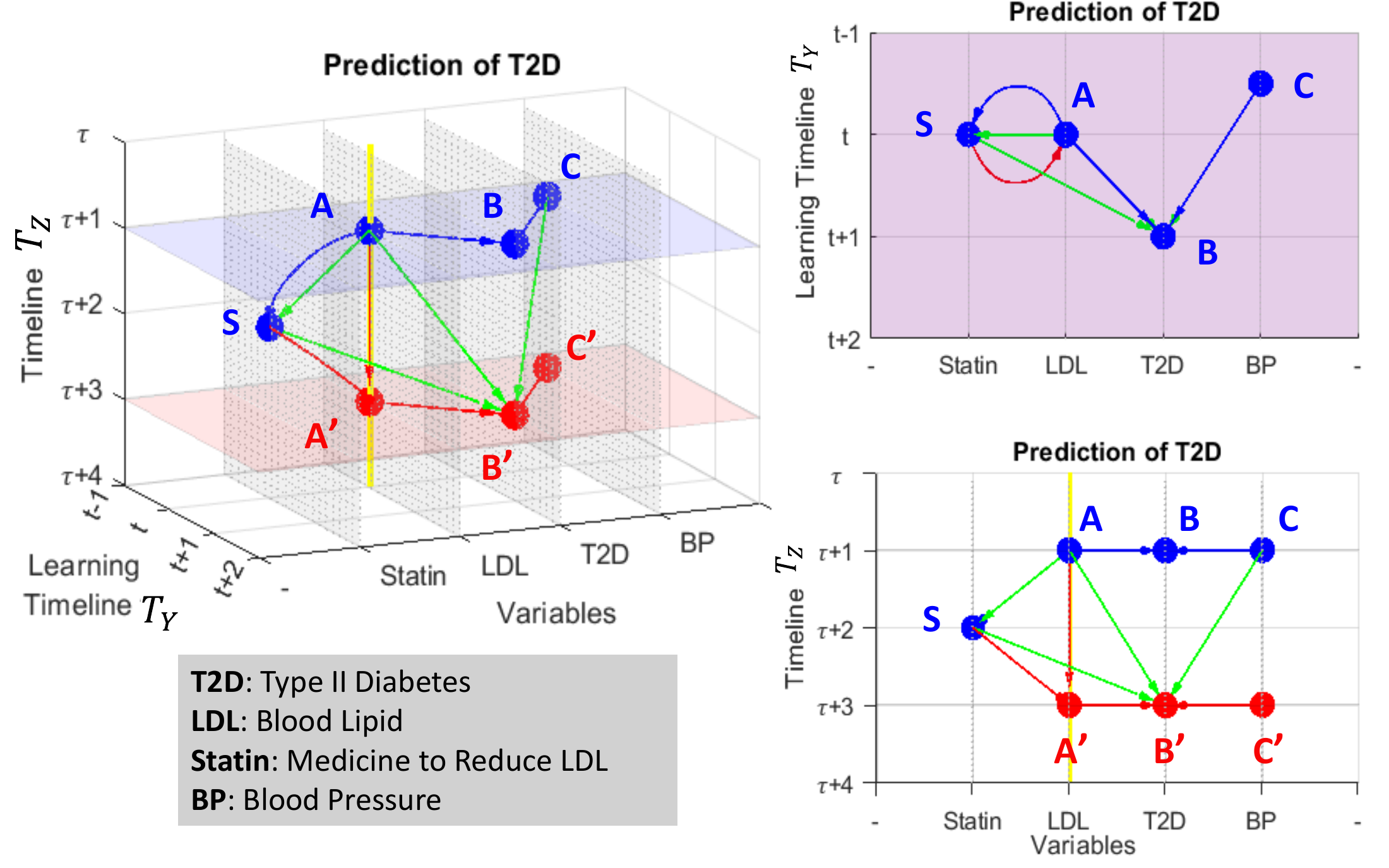}
\vspace{-4mm}
\caption{A 3-dimensional do-DAG with two timelines. The target is to predict the risk of developing T2D given values of BP and LDL, which can be causally impacted by using Statin. (Curving $A \rightarrow S$ and $S \rightarrow A'$ is to avoid overlapping.)
}
\label{fig:3d}
\vspace{-2mm}
\end{figure}

Figure~\ref{fig:3d} illustrates a learning space example comprising two timelines $T_Y$ and $T_Z$, which are relatively independent because of identifying two different effects: 1) the natural progress of the disease along $T_Y$, and 2) the effects of using medicine $S$ along $T_Z$. The uniform time stamps $(\ldots, t,t+1,\ldots)$ and $(\ldots,\tau,\tau+1,\ldots)$ are called ``time-steps'' for simplicity. Their step lengths $\Delta_{t}$ and $\Delta_{\tau}$ are unrelated and can be either equal or unequal. E.g., $\Delta_{\tau}$ spans 30 days to achieve $S$'s expected full effect, while $t$ can follow the calendar with $\Delta_{t}=$ a year, a month, or a week, etc.

Each edge has an effect function $f(\cdot)$. For individual $P$, its personalized effect can be represented as a linear transformation of $f(\cdot)$, visualized as stretching this edge for changing its time-passing speed (longer = slower). Apparently, the do-DAG $G$ can be considered the general causal knowledge of a population if and only if, for any individual $P$ in it, $P$'s personalized do-DAG is a linear transformation of $G$. So we have:

\vspace{2mm}
\noindent
\emph{\textbf{Theorem 1.} In the do-DAG space, valid SEMs must make the Markov Decision Process (MDP) eligible for $G$.}
\vspace{2mm}

\noindent
\emph{\textbf{Corollary 1.} If $\ G$ only involves a single timeline, then the regular SEMs that build up correlations are valid for $G$.}
\vspace{2mm}

\noindent
\emph{\textbf{Corollary 2.} If $\ G$ crosses multi-timelines but not contains any confounding relations, the regular SEMs are also valid.}
\vspace{2mm}

Theorem 1 can be more familiar as the Markov assumption, commonly implied to be true by default in regular SEMs. MDP is the strategy of \emph{reinforcement learning}, whose applications usually have nothing to do with multi-timelines, e.g., playing a Go game. In other words, it is rarely noticed that confounding across multi-timelines may lead to invalid Markov processing.
%and we name the bias it brings to the model as \underline{\emph{Causal Representation Bias}} (CRB).
%Since two edges define a 2D plane, in Corollary 1, the graph can define a hyperplane valid for learning.

In Figure~\ref{fig:3d}, the use of Statin is decided by doctors based on LDL criteria ($A \rightarrow S$) and will directly reduce LDL values ($S \rightarrow A'$) and subsequentially decrease the T2D risk ($A' \rightarrow B'$). The intuitive SEM to predict T2D is $B_{t+1}=f(A_t, S_t, C_t)$, intend to realize the learning along $T_Y$, as shown in Figure~\ref{fig:3d} upper-right subplot.
However, such a plane only carries the causation among $\{A, B, C\}$, where $S_t$ is not meaningful and forms a circle $A \leftrightarrow S$. The actual modeling edges are marked as green. But can $B'=f(A, S, C)$ be a valid SEM?
Apparently, modeling correlations among these nodes has to fix the time spans of all edges in the confounding ($\overrightarrow{AS}$, $\overrightarrow{SB'}$, and $\overrightarrow{AB'}$) simultaneously. Considering each individual may have a unique relative time-passing speed between $T_Y$ and $T_Z$, differentiated from others, this SEM can hardly guarantee validity for all individuals.

%\vspace{-2.mm}
\subsection{Individualization and Generalization}
Suppose in Figure~\ref{fig:3d}, all data are from one patient, i.e., only one individual exists, then the validity can be naturally confirmed. So, the violation of Theorom 1 has three necessary conditions.

\vspace{2mm}
\noindent
\emph{\textbf{Definition 3.} The modeling bias brought by invalid SEMs is named \underline{Causal Representation Bias} (CRB), formed by 3 factors: 1) confounding, 2) across multi-timelines, 3) individual-level heterogeneity.}
\vspace{2mm}

Notably, \emph{being heterogeneous} is not equal to, but more restrictive than \emph{not i.i.d}. For example, if patient $P_1$ is half slower than $P_2$ on all the edges of do-DAG, then collected data from them are not i.i.d, but they cannot be counted as heterogeneous, since they have the same relative time-passing speeds among all independent timelines.

% Interestingly, CRB comes from the heterogeneous \emph{individual-level} differences in cohort data, e.g., one cannot infer that $P_1$ is always $0.6$ times faster than $P_2$ for any medical effects just based on $M_1$.
% In health informatics, assuming no CRB commonly means all samples are from the same patient who runs a consistent causal system in the body, thus, whose underlying causal graph $G$ can be modeled in one-time learning, i.e., the valid learning plane carrying $G$ exists. 
%However, it is still possible that individual-level features are not heterogenous but linearly differ in some other domains' questions.

% \vspace{2mm}
% \noindent
% \emph{\textbf{Corollary 2.} The common causal knowledge $G$ of a cohort cannot be none-CRB modeled in one-time learning unless all \underline{individualized} causality can be represented as linear transformations of $G$.}
% \vspace{2mm}

Indeed, collecting EHR(Electric Healthcare Records) data from one patient as many as enough to build a model is almost impossible in the healthcare domain. So in practice, learning patients' individual-level features is very challenging.
Speaking of big heterogeneous data, Machine Learning can build great models, but due to involved CRBs, they are usually not generalizable.

Recently, Scholkopf et al. \cite{scholkopf2021toward} specified the generalization problem of causal learning and first raised the term \emph{Causal Representation}. They point out that ``most work in causality starts from the premise that the causal variables are given'', and summarize the challenges we are facing on the way to realizing AI:
1) The weak robustness of modeling; 2) No mechanisms to make the learned causal knowledge reusable; 3) No universal way to model general causal knowledge from heterogeneous data.

A typical generalization problem is described in experiments of Section 4, coming from the hydrology area: Since the causation that environmental conditions influence streamflow observations is based on the same physical regulations on the earth, how can we build SEMs that can be generalized to any watershed?
In this question, an individual is a watershed, like a patient in EHR data.
The only difference is that the former can individually produce enough data samples as needed, but the latter cannot.
So they both face the difficulty of dealing with individual-level heterogeneities but for different purposes.

\vspace{2mm}
\noindent
\emph{\textbf{Definition 4.} For data with $n$ samples coming from $m$ individuals, let do-DAG $G$ be their general causal knowledge, then \underline{individualization} means finding $G_j$ compared to $G_i$ or $G$, and \underline{generalization} means finding $G_j$ by given $G_i$ or $G$, where $i \neq j \in 1,\ldots,m $. }
\vspace{2mm}

\begin{figure}[h!]
\vspace{-3.5mm}
\hspace*{-4mm}
\includegraphics[scale=0.46]{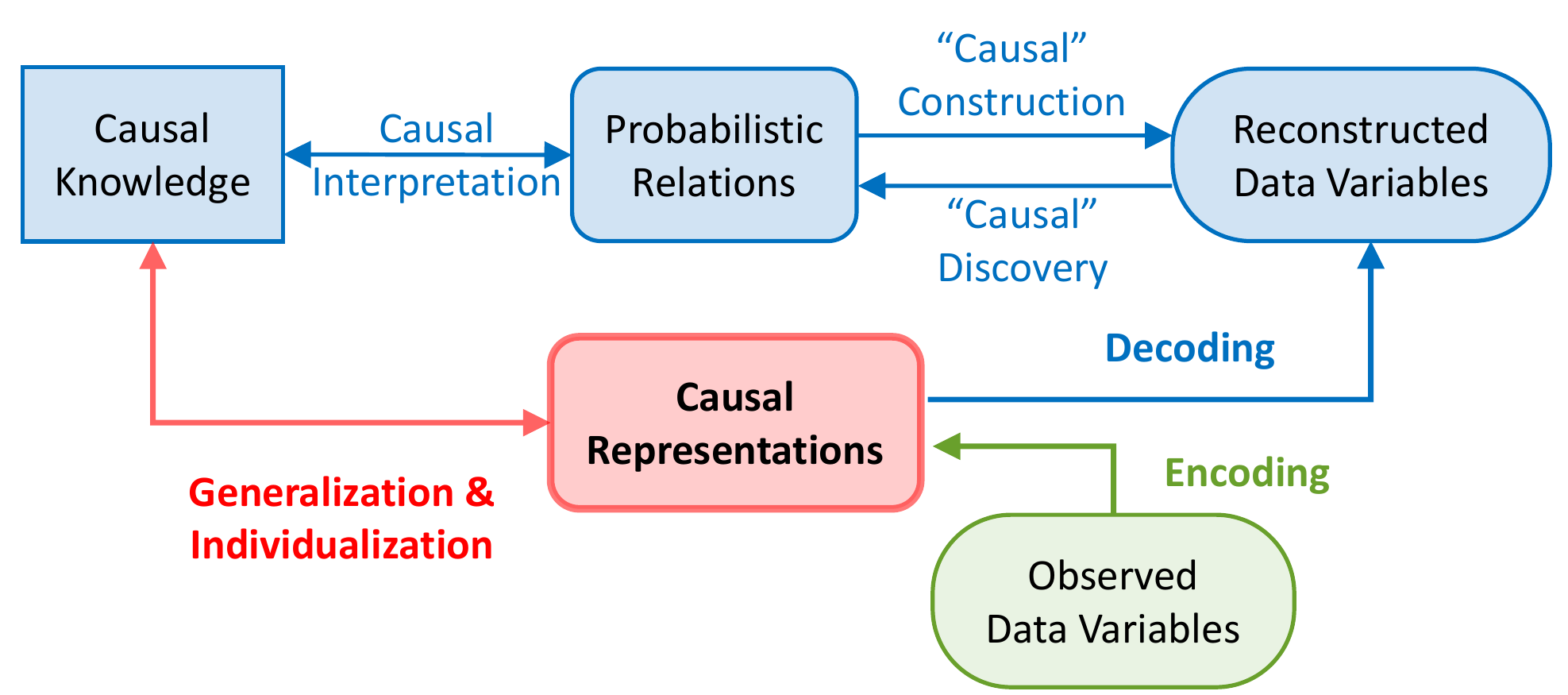}
\vspace{-5mm}
\caption{Proposed framework: Causal Representation Learning (CRL). \textcolor{red}{Red} = the causal latent representation space; \textcolor{green}{Green} = the original data space; \textcolor{blue}{Blue} = the reconstructed data space. Here ``causal'' means they are indeed correlations.}
\label{fig:new}
\vspace{-5mm}
\end{figure}

\vspace{-2.mm}
\subsection{Causal Representation Learning (CRL)}
In Machine Learning, leveraging domain knowledge to improve causal models is to deal with potential CRBs case by case. But one can hardly be sure if individuals are heterogeneous, which effects share the same timeline, or whether unknown confounding exists.
So, conservatively, we should model the causal effects for each pair-wise causation, i.e., each edge, until going through $G$. 
%causal construction means modeling known causations, like estimation of causal effects between specified variables.

Figure~\ref{fig:new} shows the proposed generic framework CRL. All causal effects are modeled in the latent space. To model $A\rightarrow B$, we first extract representation vectors for $A$ and $B$, respectively, and then build an RNN model between them. The graphical structure of $G$ will be constructed sequentially by ``stacking'' these established effects models and performing adjustments accordingly. E.g., to stack $A\rightarrow B$ with $B\rightarrow C$, the value of $B$'s representation vector needs to be modified accordingly, but simultaneously, it can stay the same representative to the observed values of $B$. Since latent vectors can capture an individual's status (instead of correlation over a specific time), do-DAG can be realized in latent space. Such a space indeed plays the role of ``the axis of different nodes $\mathcal{V}$''.

% Under the conventional framework, causal learning is restricted in observational data space based on unmodifiable variable values. 
% Further, considering the black-box nature of neural networks, it is reasonable for many DL works to assume data variables adequately represent underlying relations, to avoid further demand for interpretation.
% And the advantage of DL in handling numerous variables on large-scale data can make CRBs easier to be overlooked.

The term ``CRL'' was initially raised by Scholkopf et al.\cite{scholkopf2021toward}, who mainly focus on computer vision applications, not intending to redefine concepts. Their ``disentangled representation'' method is similar to our strategy and also aims to avoid potential CRB.

% In such a process, the feature vectors representing data variables can have values changing but without loss of representativeness.
% Reconstruction via autoencoders (encoding + decoding) is to realize the nonlinear transformations between the latent learning plane and the original data space; but differently, the reconstructed variables are friendly to one-time learning without concerns about potential CRBs.

% Previously, there may not be a strong demand for CRL, because leveraging prior knowledge to calibrate models case-by-case can commonly satisfy the requirement.
% But nowadays, with the information explosion, statistical models are approximately reaching their capability ceiling, and machine learning has to face larger-scale questions much more than before.

\vspace{2mm}
\noindent
\emph{\textbf{Theorem 2.} Let data matrix $X$ augmented by column vectors of observed values of $\mathcal{V}$ in $G$, then to adequately representing $G$, the latent space must have a dimensionality $\geq rank(X)$.}
\vspace{2mm}

Theorem 2 is derived from the principle that autoencoders learn the subspace spanned by the top principal components of $X$ \cite{baldi1989neural, elad2018from, wang2016auto}, more well-known as PCA. %(principal component analysis).
Unlike the regular autoencoder for lower-dimensional feature extraction, this means we must design the higher-dimensional one from scratch (introduced in Section 3).

Furthermore, we hypothesize that reducing the latent space's dimensionality lower than $rank(X)$ can make the causal relations in $G$ more obvious (but less adequate).
It is inspired by the principle that autoencoders can produce aligned latent spaces by stretching along the top singular vectors of $X$ \cite{jain2021mechanism}, which has been widely used in NLP
%(natural language processing) 
for identifying word analogies \cite{pennington2014glove, rong2014word2vec}, and in bioinformatics to embedding gene expression\cite{belyaeva2021causal, lotfollahi2019scgen}. But it requires more corroboration;
once verified, maybe CRL can help to improve causal inference's efficiency.

\vspace{-1mm}
\section{Related Work}\label{sec:relate}

\subsection{Statistical Causal Learning}
Conventional causal learning usually aims to learn causal DAGs as Bayesian networks (BNs) by estimating conditional independencies among observed variables. In classical statistics, improvements are mainly in two directions. One is on methods of determining independencies, like the constraint-based (e.g., Fast Causal Inference and PC) and score-based (e.g., Greedy Equivalence Search) ones \cite{scheines1997introduction}. Another is inferring the connection between discovered dependencies and causal structures\cite{glymour2019review}. 
In machine learning, the graphical models, often presented as Structural Equation Models (SEMs) or Functional Causal Models (FCMs) \cite{glymour2019review, elwert2013graphical}, are
commonly used to leverage causal knowledge.
But they are still based on probabilistic dependencies without identifying causality from correlation.

In regular SEMs modeling correlations, the defined CRB is usually reflected as not-fully observed confounding. Thus Sufficiency assumption (i.e., no hidden confounder) is critical to decide\cite{scheines1997introduction}. On the other hand, statistics values models' linearity to stay interpretable, which further confuses CRB and confounding bias. E.g., in Figure~\ref{fig:3d}, for both $A'=f(S, A)$ and $B'=f(S, A)$ linear SEMs, backdoor adjustment\cite{pearl2012calculus} is always preferred to de-confound the association between $\{A, S\}$, although the former has no CRB problem.

In our opinion, \emph{timeline} is the key to distinguishing causality from correlation. In many
causal questions, time is only a parameter of correlation, e.g., ``Does a higher smoking rate cause more lung cancers?'' It can be answered in a $5$, $10$, or $30$ period but makes no difference in modeling. However, the causal effects of interventions can be time series, like to answer ``How does this medicine influence lungs?'' timeline is indeed an expected describer of the effects.

% They assume the value of $Y$ to be a deterministic function of its direct causes $X$ with noise term $\epsilon$, $Y=f(X,\epsilon;\theta)$ where $\theta$ indicates the parameters set involved in $f$.
% %, which is subject to the constraint of pairing a directed graph with a joint probability distribution. 
% They essentially aim to decompose a causal DAG into a product of more superficial vectors, where the separation is by conditional independence.
% It has been shown that they can distinguish different DAGs from an equivalence class and are good at leveraging prior knowledge \cite{glymour2019review}.

\begin{figure}[h!]
\vspace{-2mm}
%\centering
\hspace*{-8.5mm}
\includegraphics[scale=0.44]{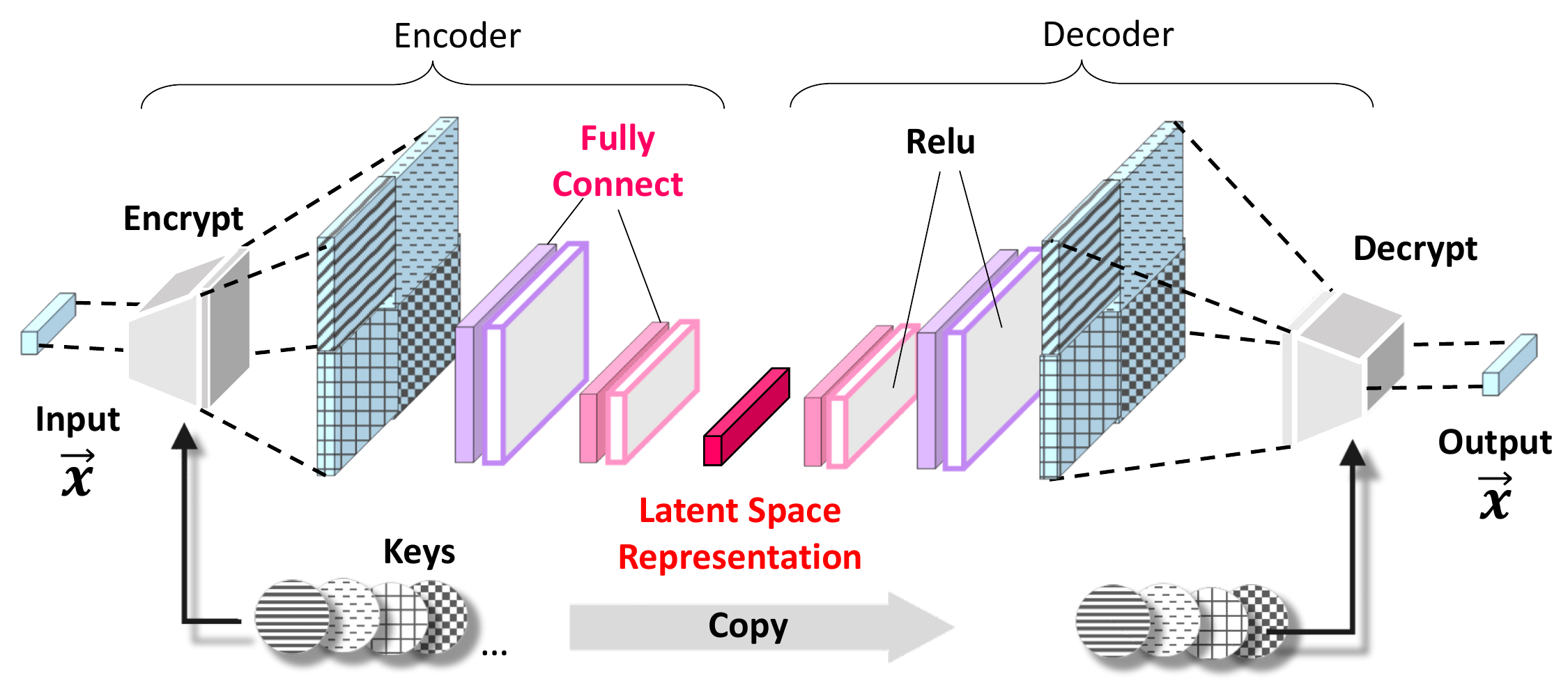}
\vspace{-7mm}
\caption{Proposed architecture of the higher-dimensional representation extraction autoencoder.}
\label{fig:arch}
\vspace{-3mm}
\end{figure}

%\vspace{-5.9mm}
\subsection{Deep Learning in Causality}
\label{sec:relate_1}

% The concept of representation learning was proposed about 100 years ago, including the unsupervised principal component analysis (PCA) in 1901 and supervised linear discriminant analysis (LDA) in 1936.
% In recent years, the deep network architectures for representation learning have turned out to be widely applied \cite{zhong2016overview}.
% Data representation describes how much the explanatory vectors of variation behind the data have been revealed \cite{bengio2012unsupervised}, and thus plays an important role in determining the success of machine learning methods. 
% In the quest for applicable Artificial Intelligence (AI), more powerful methods are motivated to be designed in this direction.

Deep Learning (DL) can handle non-linear and higher-order dependencies among variables and simultaneously provide global solutions, so it is trending to leverage its effectiveness and the interpretability of causal inference\cite{luo2020causal}.
A significant contribution is converting the discrete combinational constraints on DAG's ``acyclic-ness'' into continuous ones to realize the global optimization on networking\cite{zheng2018dags, lachapelle2019gradient}, and has been further generalized into nonparametric modeling \cite{zheng2020causal}.
To achieve interpretability, some works attempt to reconstruct causal knowledge by properly designing architecture\cite{ma2018using}, while others try to infer dependencies directly from neuron weights\cite{lachapelle2019gradient, zheng2020causal}. 
Such neural architecture search (NAS) methodology aims to realize accuracy and transparency simultaneously but is still far from general success\cite{luo2020causal}. 

Unlike statistical modeling, where the intention of removing confounding can help prevent CRB, DL's black-box nature and global optimization strategy make the inherently existing CRBs more challenging to be found. We implicitly hope DL can automatically solve the confounding biases concerning causal inference, but it shows to be unrealistic\cite{li2020teaching}.
People barely notice that, in our awareness, causal DAGs used to be 2-dimensional because we have an intuitive understanding of timelines. But AI takes all relationships as associations, so timelines must be explicitly constructed; otherwise, the individuals' heterogeneity can hardly be described.

%\vspace{2mm}
\section{Methodology of Causal Representation}
\label{sec:method}

From the probabilistic perspective, causal representation learning (CRL) is like to ``simulate'' the dependencies among variables in a latent space.
All observed data variables will be individually converted as representation vectors, then the spaces between them allow us to build the effect models.
Suppose observed value $x$ has a corresponding latent vector $h$, their joint distribution can be decomposed as $P(x,h)=P(x|h)P(h)$, where $P(h)$ is a \emph{prior} and $P(x|h)$ is a \emph{likelihood}. Thus a general latent model can be inherently interpreted as a latent cause model \cite{bengio2012unsupervised} directed from $h$ to $x$, denoted as $h\rightarrow x$. This forms the theoretical basis of realizing CRL.

Consequently, we can reasonably decomposite the causation from observation $x$ to $y$ (denoted as $x\Rightarrow y$) as a causal chain $x\rightarrow h\rightarrow v\rightarrow y$ where $v$ is the corresponding latent vector of $y$, and $x\rightarrow h$ relies on a \emph{posterior} $P(h|x)$. Such that the dependency between $x$ and $y$ is represented as the \emph{posterior} distribution $P(v|h)$. Suppose we have the parametrized model $v=f(h)$, then tuple $(v,h,f)$ can completely represent the causation $x\Rightarrow y$ in latent space. To be convenient, we call %$v$ and $h$ as \emph{Latent Variables} and
$f$ as the \emph{Latent Causal Effect}.

The ultimate goal of CRL is to build up SEMs in the latent space by ``stacking'' these posterior distributions according to demands.
%which are flexible to be generalized or individualized. 
For example, given causation $x\Rightarrow y \Rightarrow z$, we are interested in the indirect causal effect of $x$ on $z$ via $y$, decomposed as $x\rightarrow h\rightarrow v\rightarrow t\rightarrow z$, where $t$ is the latent vector of $z$, then what we persue is $P(t|h)=P(t|v)\big|_{v=f(h)}$. Given $f$ is ready, we only need to learn the latent causal effect $g$, such that $t=g(v)\big|_{v=f(h)}$ by stacking the unknown \emph{posterior} $P(t|v)$ onto the known \emph{posterior} $P(v|h)$.

%A Causal Representation of pairwise causation consists of representations of the two Variables and a learned model representing the causal Effect between them.
%Furthermore, it is required that the \emph{posterior} distributions in such Causal Representations have to be ``stackable'' to implement reconstruction and the discovery of graphical causal structures in latent space.
In the following, we will start from the proposed autoencoder architecture to realize the required higher-dimensional representation extraction (\ref{sec:method_1}) along with the initial design of its critical layers (\ref{sec:method_2}); then, introduce how to model the latent causal effect and stack them to complete a graphical structure (\ref{sec:method_3}); and lastly demonstrate a causal discovery algorithm in latent space (\ref{sec:method_4}).

\subsection{Autoencoder Architecture}
\label{sec:method_1}

Regular autoencoders are commonly applied on large-dimensional objects (like images with numerous pixels) and extract the lower-dimensional features accordingly, which is more like a distillation process. However, our appeal is to individually convert the causal variables into a higher dimensional latent space as representation vectors and model their dependencies (i.e., latent causal effects). 
Considering the do-DAG's complexity, this latent space should provide sufficient freedom for all possibly needed causal modeling.
Theorem 2 gives the lowest dimensionality choice for adequate representation, but we temporally shelve further discussions about possible boundaries and manually choose one for our experiments. We have $10$ variables whose lengths are between $1\sim 5$ and summarized to be $32$ in total (refer to Table~\ref{tab:tower}). We initially chose $64$-dimensional latent space, but it was highly redundant since the variables' associations are pretty strong; then, we gradually reduced it to $16$-dimentional.

\begin{figure}[h!]
\vspace{-1mm}
\includegraphics[scale=0.34]{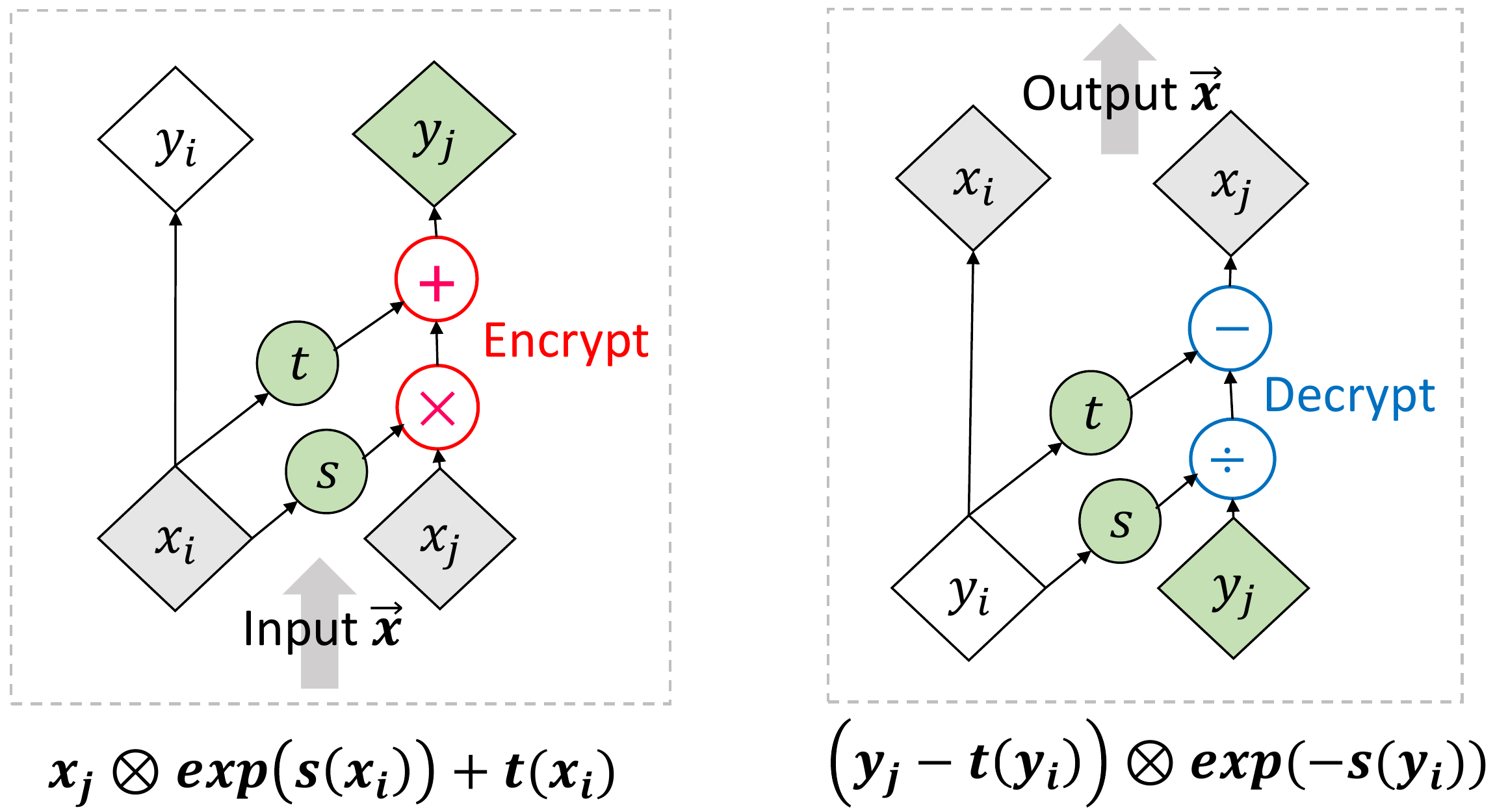}
\vspace{-3mm}
\caption{Encrypt and Decrypt with \emph{Key} $\mathbf{\theta}$ specifying $s$ and $t$.}
\label{fig:extractor}
\vspace{-3mm}
\end{figure}

\vspace*{-2mm}
\begin{figure}[h!]
\centering
\includegraphics[scale=0.4]{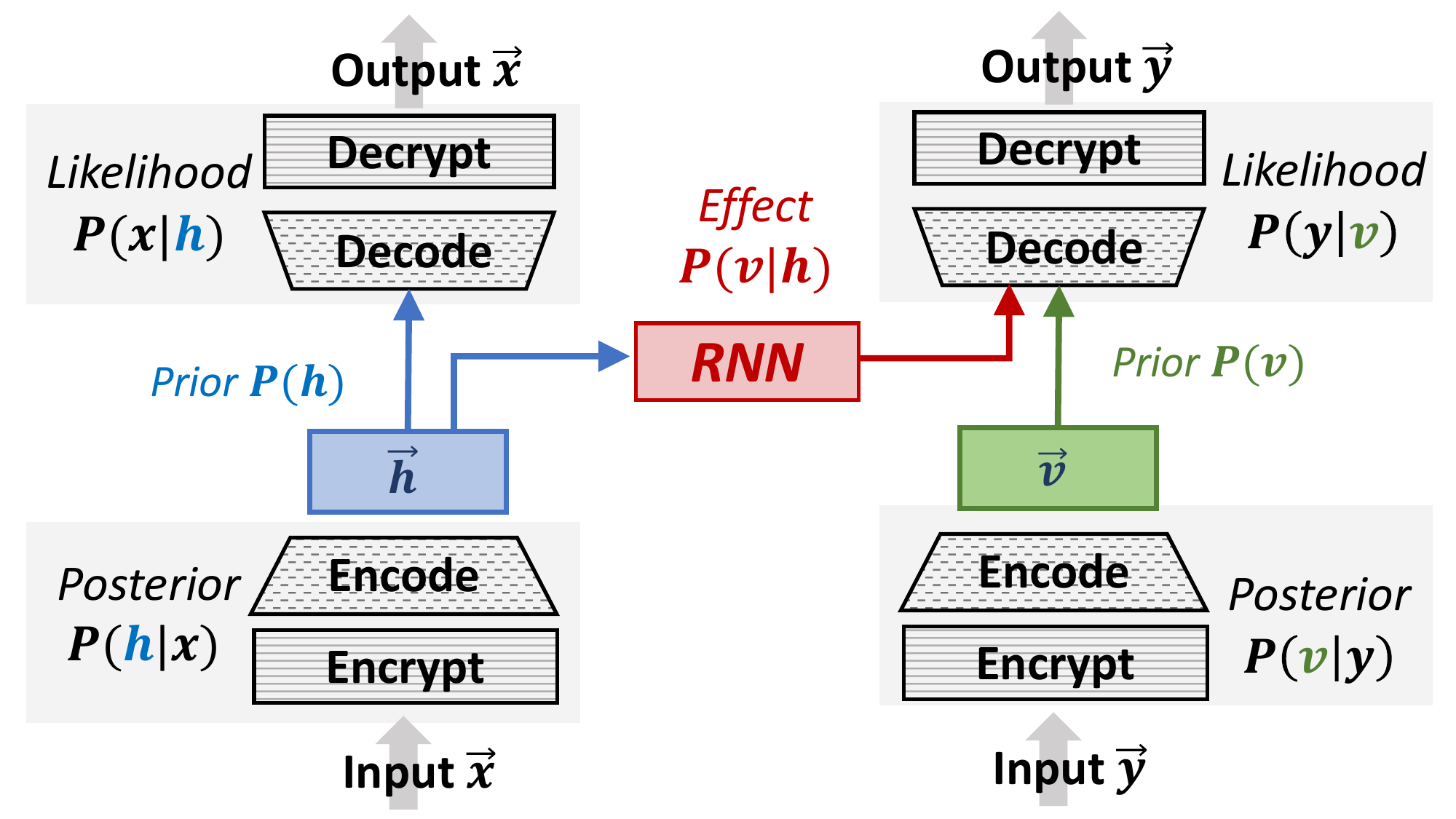}
\vspace{-3mm}
\caption{Architecture of learning the latent causal effect $h\rightarrow v$ as  a RNN model.}
\label{fig:bridge}
\vspace{-2mm}
\end{figure}

For better reconstruction, the input variables need an extension, where the extended one can sufficiently encode the associative information among its values. More importantly, this process has to be \emph{Invertible} to achieve high accuracy for the output layer, unlike in regular autoencoders, little concern is about the accuracy of recovering each pixel.
We design a pair of symmetric layers, \emph{Encrypt} and \emph{Decrypt}, respectively located at the Input and Output of the autoencoder, as shown in  Figure~\ref{fig:arch}.

The Encrypt works as a ``feature amplifier'' for extracting higher-order features from the input vector, while Decrypt exactly reverses this process.
How much higher order should be reached is on practical demand. For example, a \emph{double-wise} amplifying is to encode associations between every two digits, and \emph{triple-wise} encodes every three digits.
In this work, we only use the double-wise one.
Specifically, all possible combinations of two digits from this vector will be ``encrypted'' using a set of constants named \emph{Key}, generated by Encoder.
Then the \emph{Key} will be copied by the Decoder to ``decrypt'' its output back to two digits.
For every digits-pair, one \emph{Key} means one-time amplifying, so with multiple \emph{Keys}, we can get a group of vectors differently amplified from the same input digits; then concatenate them to form a much longer new vector to be their double-wise amplified input.
For an input vector with length $n$, one-time double-wise amplifying through all digits-pairs can form a $(n-1)*(n-1)$ sized matrix, so we use a square (not implying a 2-dimensional vector) to represent it in Figure~\ref{fig:arch}.
This occasion has four \emph{Keys}, and we have four squares with different ``signatures'' accordingly.
As a metaphor, the role of \emph{Key} here is the same as the filter in regular autoencoders.

\subsection{Vector Encrypt and Decrypt}
\label{sec:method_2}
Let $(x_1, x_2, \ldots, x_n)$ denote the input vector with length $n$.
Then Encrypt can be defined as a function $f(x_j ; x_i,\theta)$, seen as a transformation of $x_j$ using $x_i$ as weights and the \emph{Key} $\theta$ as parameters, where $i\neq j \in 1,\ldots,n$.

Suppose $\theta=(w_s,w_t)$ specifies two functions, $s(x) = w_s x$ and $t(x) = w_t x$, and $f(x_j ; x_i,\theta) =  x_j \otimes exp(s(x_i)) + t(x_i)$,
where $\otimes$ denotes the element-wise multiplication. Let $y_j = f(x_j ; x_i,\theta)$ represent corresponding output of $x_j$.
Accordingly, the inversed function $f^{-1}$ in Decrypt will be $(y_j-t(y_i)) \otimes exp(-s(y_i))$. Figure~\ref{fig:extractor} illustrates these two symmetric processes. 
Since calculating $f^{-1}$ does not involve inversed function $s^{-1}$ or $t^{-1}$, the two elementary functions $s$ and $t$ can be very flexibly defined as either linear or non-linear.

Let's assemble all $f$ functions as $\mathcal{F}(X; \Theta)$, where $\Theta$ is the set of all $\theta$.
Then the two layers, Encrypt and Decrypt, can be denoted as $Y=\mathcal{F}(X; \Theta)$, and $X = \mathcal{F}^{-1}(Y; \Theta)$ respectively.
Source code\footnote
{https://github.com/kflijia/bijective\_crossing\_functions/blob/main/code\_bicross\_extracter.py} is provided as in a complete demo.
It is worth highlighting that the work of Dinh, L. et al. inspires the design proposed above \cite{dinh2016density}. 
One can perform higher-order encoded associations using the same principle as a possible extension.

\subsection{Latent Causal Effects Stacking}
\label{sec:method_3}

Figure \ref{fig:bridge} displays the architecture of learning the latent causal effect $h\rightarrow v$, representing the observational causation $x\Rightarrow y$, which is indeed to construct the causal chain $x\rightarrow h\rightarrow v\rightarrow y$.
The latent space vectors $h$ and $v$ are established representations for $x$ and $y$, respectively, the same dimensional ($h,v\in \mathcal{R}^{L}$) but independent of each other.
In the learning process, $h$ and $v$ will keep updated during optimizations to get ``closer'' to each other.
It can be seen that both $h$ and $v$ are shifting in $\mathcal{R}^{L}$ space toward finding the proper locations minimizing the distance between them and, at the same time, letting the $RNN$ sufficiently model their dependence $P(v|h)$, which represents the estimated causal effect $h\rightarrow v$ in $\mathcal{R}^{L}$.

\vspace{1mm}
There are three optimizations with three different objective functions, sequentially executed in each optimization iteration: 
\begin{enumerate}
    \item {Update \emph{posterior} $P(h|x)$ and \emph{likelihood} $P(x|h)$ to minimize the reconstruction error of $x$}.
    \item {Update \emph{posterior} $P(h|x)$ and latent causal effect \emph{posterior} $P(v|h)$ to minimize the reconstruction error of $y$}.
    \item {Update \emph{posterior} $P(v|y)$ and \emph{likelihood} $P(y|v)$ to minimize the reconstruction error of $y$}
\end{enumerate}
%\vspace{1mm}

%\vspace{-2.7mm}
\begin{figure}[h!]
\centering
\includegraphics[scale=0.45]{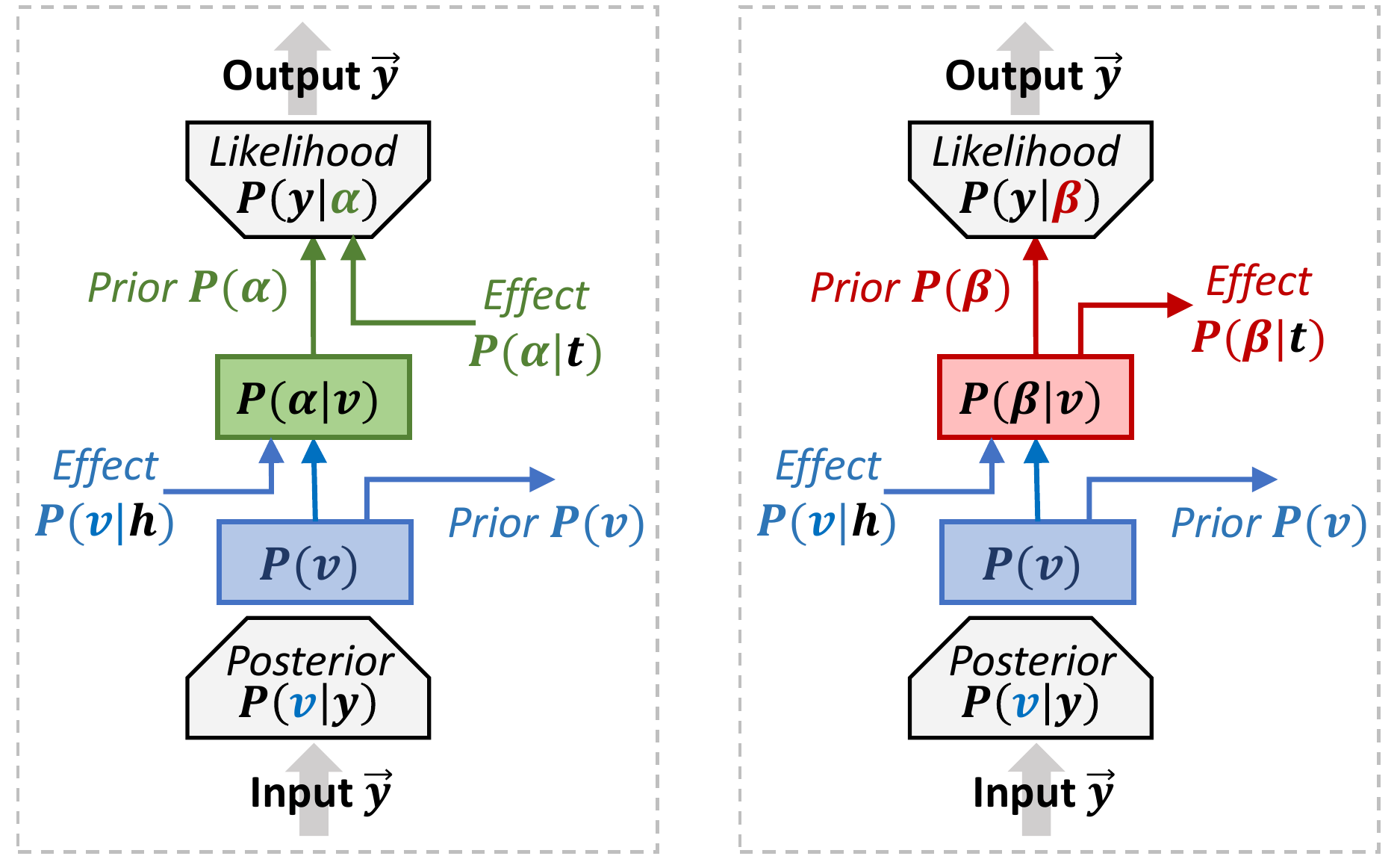}
\vspace{-2mm}
\caption{Example of stacking two pair-wise causations to represent the complete causal chain over $x,y$ and $z$.}
\label{fig:stack}
\vspace{-4mm}
\end{figure}

%\vspace{-8mm}
Specifically, ``stacking'' two causal effects in the latent representation space means that given one is well established, we aim to build up the second one starting from either end of the first one to ultimately establish a 3-node causal relationship.
%chain, which can eventually shape as a colliding, a confounding, or a chain according to needs.
This is essentially equal to stack two \emph{posteriors}, e.g., stack $P(t|v)$ onto $P(v|h)$ to form the chain $h\rightarrow v\rightarrow t$.
There are four possible occasions by combining two binary options: is the \emph{prior} at the bottom a \emph{cause} or a \emph{result}? And how will the \emph{prior} at the top be?

Suppose variable $y$ is the connecting point where the stacking will happen, and we have $x$ and $z$ at each side of $y$ to be the candidates for connection, represented as vectors $h$ and $t$. 
As shown at Figure~\ref{fig:stack} left side, the \emph{prior} $P(v)$ at the bottom can be either a \emph{cause} to the other one (output toward the right) or
the \emph{result} caused by $h$ (input as causal effect $P(v|h)$). 
While for the top layer, as shown at Figure~\ref{fig:stack}, there are two possible roles of $t$: to be the new \emph{cause} that inputs the effect $P(\alpha|t)$ (green colored at left), or the new \emph{result} that accepts the effect $P(\beta|t)$ (red colored at right).
In the optimization process of the top layer, $t$ always represents the fixed $prior$ to support updating the \emph{posterior}, either $P(\alpha|t)$ or $P(\beta|t)$.

Multiple combinations of inputs and outputs exist for each occasion. The selection should be based on the practical demand of causal constructions (i.e., establish a known causal relationship) or discoveries.
For example, let $\mapsto$ be the notation that connects entrance and exit, then,
$P(v|h) \mapsto y$ means output $P(y|x)$, 
$P(\alpha|t) \mapsto y$ means output $P(y|z)$, and $y \mapsto P(\beta|t)$ means output $P(z|y)$ with the ground truth $y$ input, while
$P(v|h) \mapsto P(\beta|t)$ means output $P(z|y)\big|_{y=f(x)}$ where $f$ is the estimated causal effect $x\Rightarrow y$.

From a geometric perspective, a proper latent representation space is gradually formed
along with every successful stacking in $\mathcal{R}^{L}$. 
Suppose dimensionality $L$ is large enough to provide sufficient freedom for going through the do-DAG $G$. In that case, we can finally have a group of representation vectors optimized as distributed tightly as possible in $\mathcal{R}^{L}$.

\subsection{Discovery Algorithm in Latent Space}
\label{sec:method_4}

%\vspace{-3mm}
\begin{figure}[h!]
\centering
\includegraphics[scale=0.38]{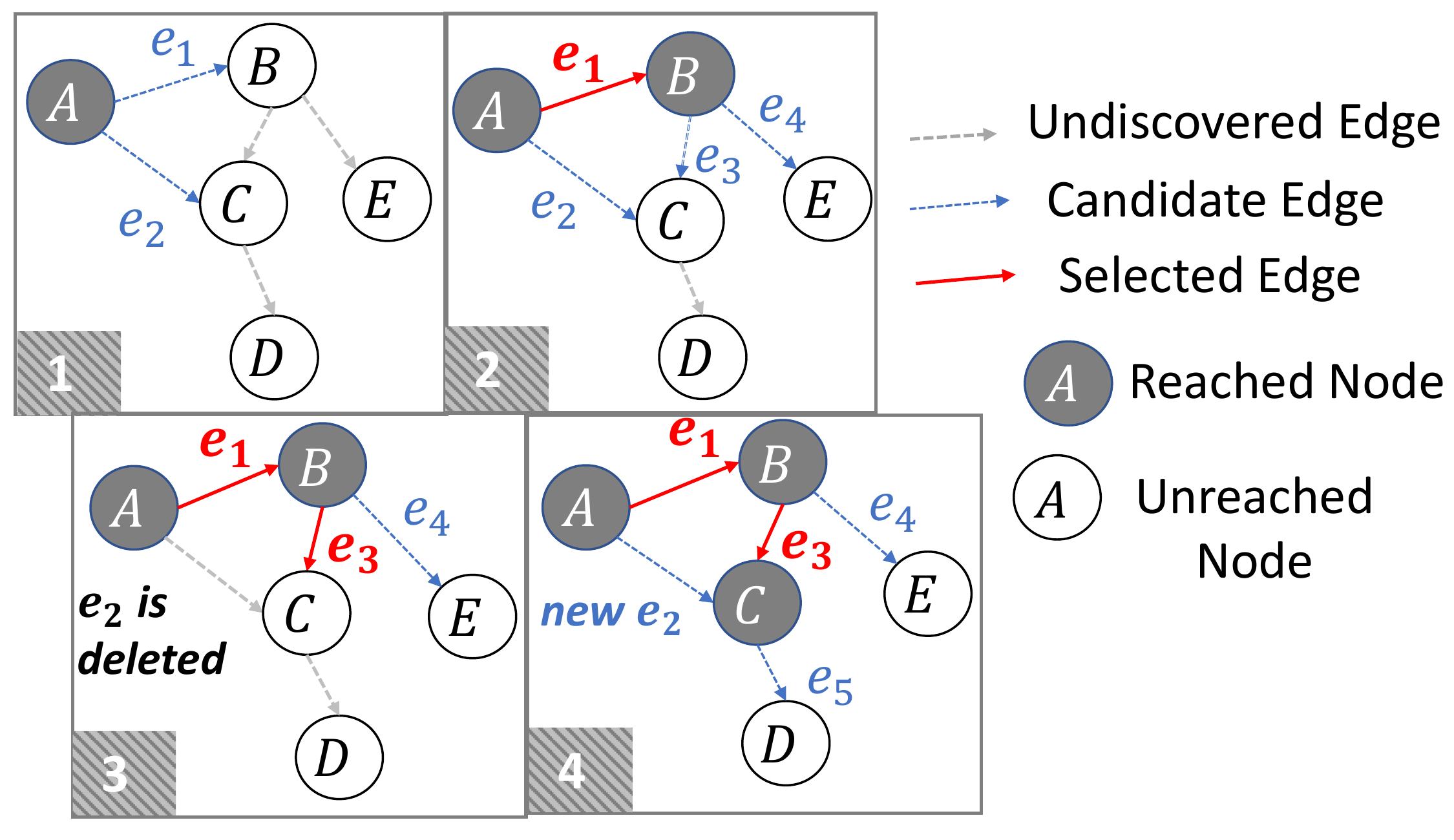}
\vspace{-2mm}
\caption{Example of Causal Discovery in the latent representation space. Nodes indicate representation vectors.}
\label{fig:discover}
\vspace{-3mm}
\end{figure}

Here we adopt the heuristics strategy to realize causal discovery in latent space among the well-established representation vectors; and use KLD (KL-Divergence) as the metric to evaluate the similarity between them in $\mathcal{R}^{L}$, as shown in Algorithm 1.
Another consideration is adopting the conventional MSE evaluation. However, it has been demonstrated that the MSE-based optimization subject to acyclicity constraint has some inherent restrictions, such that the model may be misled and eventually dominated by the data variances instead of the underlying causal relations \cite{reisach2021beware,kaiser2021unsuitability}.

An illustrative example of the latent space causal discovery is shown in Figure~\ref{fig:discover}. 
In these four consecutive steps, edge $e_1$ and edge $e_3$ are selected one after another. The selection of $e_1$ makes node B reachable as the start node of $e_3$.
At step $3$, the learned causal effect of $e_2$, from $A$ to $C$, is deleted from the candidate edges and needs to be recalculated because edge $e_3$ has reached $C$ and changed current causal conditions on $C$.

\vspace{-2mm}
\begin{algorithm}
\footnotesize
\SetAlgoLined
\KwResult{ ordered edges set $\mathbf{E}=\{e_1, \ldots,e_n\}$ }
$\mathbf{E}=\{\}$ \;
$N_R = \{ n_0 \mid n_0 \in N, Parent(n_0)=\varnothing\}$ \;
\While{$N_R \subset N$}{
$\Delta = \{\}$ \;
\For{ $n \in N$ }{
    \For{$p \in Parent(n)$}{
        \If{$n \notin N_R$ and $p \in N_R$}{
            $e=(p,n)$;\\
            $\beta = \{\}$;\\ 
            \For{$r \in N_R$}{
                \If{$r \in Parent(n)$ and $r \neq p$}{
                    $\beta = \beta \cup r$}
                }
            $\delta_e = K(\beta \cup p, n) - K(\beta, n)$;\\
            $\Delta = \Delta \cup \delta_e$;
        }
    }
}
$\sigma = argmin_e(\delta_e \mid \delta_e \in \Delta)$;\\
$\mathbf{E} = \mathbf{E} \cup \sigma$;\\
$N_R = N_R \cup n_{\sigma}$;\\
}
 \caption{Latent Causal Discovery with KLD Metric}
\end{algorithm}
%\vspace{-3mm}

\begin{table}[h!]
%\caption{Algorithm Notation List }
\vspace{-6mm}
%\label{tab:notation}
  \begin{center}
    \label{tab:table1}
    \resizebox{\columnwidth}{!}{%
    \begin{tabular}{|l|l|}
    \hline
       $G=(N,E)$ & graph $G$ consists of nodes set $N$ and edges set $E$\\
       $N_R$ & the set of reachable nodes\\
       $\mathbf{E}$ & edges in order of being discovered\\
       $K(\beta, n)$ & KLD metric between the causes set $\beta$ and effect node $n$\\
       $\Delta=\{\delta_e\}$ & the set of all KLD Gain $\delta_e$ for each candidate edge $e$\\
       $n$,$p$,$r$ & notation of node\\
       $e$,$\sigma$ & notation of edge\\
      \hline
    \end{tabular} %
    }
  \end{center}
\vspace{-4mm}
\end{table}

\vspace{3mm}
\section{Feasibility Experiments}
\label{sec:experiment}
% The proposed Causal Representation Learning (CRL) has two major strengths: 
% 1) It can downgrade multi-dimensional causality to be a regular correlation (1-dimensional) to adjust inherent confounding biases (i.e., CRBs) effectively. 
% 2) It can establish \emph{generalizable} Deep Learning causal models and keep a good model interpretation.

The feasibility experiments are designed to verify that the proposed autoencoder architecture can effectively perform higher-dimensional representation extractions, and accordingly, we can establish the causal effects estimations successively in the latent space. Eventually, we aim to completely build up the graphical causal structure of do-DAG, based on which the general causal discovery and construction processes can be accurately performed.

% confirm that: 1) the proposed autoencoder with \emph{Encrypt} and \emph{Decrypt} can effectively extract higher-dimensional causal variable representations.
% 2) the causal effects can be successfully represented in latent space and used to complete graphical causal constructions.
% 3) the heuristic causal discovery can effectively work in latent space.

The adopted experimental data is a professional synthetic hydrology dataset, with the learning task of forecasting streamflow by given observed environmental conditions (e.g., temperature and precipitation). 
The purpose of performing causal representation learning (CRL) on this hydrology data is to realize causal model generalization among different watersheds, such that the causal model established on the dataset $X_1$ collected from watershed No.1, can be calibrated to fit the dataset $X_2$ from watershed No.2.
This application is motivated by sharing common knowledge across different data qualities. For example, $X_1$ is qualified enough to establish a complete graphical causal model, but $X_2$ does not.
It has been known that their underlying hydrological schemes are very similar (e.g., geographical positions are closed), but because of various unmeasurable conditions (e.g., different economic developments and land use), directly applying $X_1$'s model on $X_2$ performs poorly. 
Moreover, the existing causal models are based on physical modules with limited parameters. Thus these models' degree of freedom is far from being able to do calibrations on low-qualification data. 

Because of space limitations, we can only demonstrate the CRL modeling performances (with demo\footnote
{https://github.com/kflijia/bijective\_crossing\_functions.git}) without experimental comparisons for evaluating generalization in this paper. In future works, we plan to realize both the individualization and generalization of CRL models.
Besides, due to empirical restrictions, we are out of access to EHR data for this work; for proof of causal representation bias (CRB) in EHR, please refer to our previous work \cite{li2020teaching}.

% Our experiments are performed on a professionally calibrated synthetic hydrology dataset generated by SWAT (Soil \& Water Assessment Tool),
% a commonly used hydrology data simulation system based on physical modules.
% These physical models are jointly connected to form a causal graph.
% The question from the hydrology domain is that the models learned from nearby watersheds can hardly share their learned knowledge because existing physical methods lack generalizability.
% The next experiment on this data will be toward a practically effective model-generalization way.

%\vspace{-1mm}
\subsection{Hydrology Dataset}

SWAT (Soil \& Water Assessment Tool) is a 
professional hydrology data simulation system is commonly used to generate synthetic datasets or calibrate realistic data samples based on physical modules.
Our experiments are based on the synthetic data generated by the SWAT system, for simulating the Root River Headwater watershed in Southeast Minnesota.
From this dataset, we selected 60 successive virtual years with a daily update frequency.
The performances are mainly evaluated as the accuracies of unsupervised data reconstruction via the proposed CRL model.

The hydrology system entails water fluxes and states across the earth's space, such as snowpack melting, evapotranspiration, and soil moisture.
Its underlying causality is highly complex, exhibits nonlinearities, and contains a lot of unobservable dependencies. 
In recent years, machine learning on causal inferences naturally attracted hydrologists' attention \cite{debate_Goodwill2020}; DL-based methods are widely 
used to efficiently extract representations from time series. As a typical application, RNN models contribute state-of-the-art techniques for streamflow prediction \cite{hess-18-Kratzert}.
%\cite{hess-18-Kratzert, Kratzert2019, 2022Li_Soil}. 
Here, the word ``streamflow'' refers to a critical statement variable in hydrologic processes, observed by numerous monitors in the water body and influenced by comprehensive physical factors over months.

\begin{figure}[h!]
\centering
\includegraphics[scale=0.5]{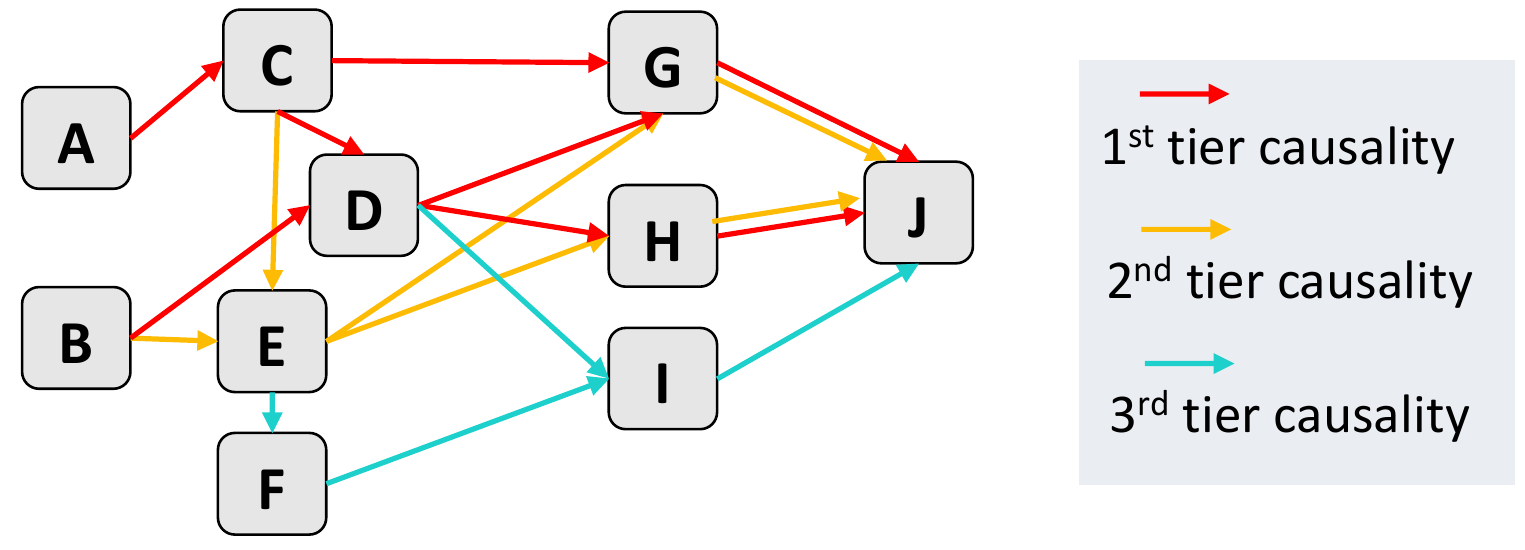}
\vspace*{-2mm}
\caption{The underlying causal graph from current hydrology knowledge. The reducing expert-expected causal strengths define the tiers of causal routines. }
\label{fig:stream}
\vspace{-2mm}
\end{figure}

Figure \ref{fig:stream} displays the ground truth graphical causal relationship used by SWAT to generate synthetic data based on current hydrology knowledge.
Explanations about the nodes are listed in Table \ref{tab:nodeID}.
The causal strengths of these routines are determined by their present contributions to output streamflow, remarked by different colors.
The hydrological interpretation is that
the surface runoff routine (1st tier causality) cause streamflow peaks more quickly than the lateral flow routine (2nd tier causality), which is more critical than the baseflow dynamics (3rd tier causality).

\vspace{-2mm}

\begin{table}[]
\caption{Explanations of the nodes in Figure~\ref{fig:stream}}
\label{tab:nodeID}
\vspace{-2.5mm}
\resizebox{\columnwidth}{!}{%
\begin{tabular}{|
>{\columncolor[HTML]{EFEFEF}}c |l|l|}
\hline
\cellcolor[HTML]{C0C0C0}\textbf{ID} & \multicolumn{1}{c|}{\cellcolor[HTML]{C0C0C0}Variable Name} & \multicolumn{1}{c|}{\cellcolor[HTML]{C0C0C0}Explanation} \\ \hline
\textbf{A} & Environmental set I & \begin{tabular}[c]{@{}l@{}}Wind Speed $+$ Humidity $+$ Temperature\end{tabular} \\ \hline
\textbf{B} & Environmental set II & \begin{tabular}[c]{@{}l@{}}Temperature + Solar Radiation \\ + Precipitation\end{tabular} \\ \hline
\textbf{C} & Evapotranspiration & Evaporation and transpiration \\ \hline
\textbf{D} & Snowpack & The winter frozen water in the ice form\\ \hline
\textbf{E} & Soil Water & Soil moisture in vadose zone\\ \hline
\textbf{F} & Aquifer & Groundwater storage\\ \hline
\textbf{G} & Surface Runoff & Flowing water over the land surface\\ \hline
\textbf{H} & Lateral & Vadose zone flow\\ \hline
\textbf{I} & Baseflow & groundwater discharge \\ \hline
\textbf{J} & Streamflow & Sensors recorded outputs \\ \hline
\end{tabular}
}
\vspace{-3mm}
\end{table}

\vspace{-2mm}
\subsection{Autoencoder Reconstruction Test}

The proposed higher-dimensional representation extraction autoencoder is used on each node in Figure~\ref{fig:stream}. Accordingly, $10$ vectors have been extracted in the latent space to represent the nodes from $A$ to $J$, respectively. 
The major challenge comes from the low dimensionality of the original data variables.
As shown in Table \ref{tab:tower}, their maximal length only reaches 5, and the prediction target, node $J$, is just a one-column variable.
We first expand each variable to have a new length of 12 by randomly repeating their columns, then augmented with 12-dimensional dummy variables of months to form 24-dimensional input variables. 
In the double-wise amplifications, every column has been extracted 23 times and augmented with itself. This way, we obtain a 576-dimensional output through the \emph{Enrypt} process. And the representation is set to be 16-dimensional.

Table \ref{tab:tower} shows each variable's statistics characteristics and the reconstruction performance (comparing the output and input of the autoencoder) evaluated by RMSE, where less RMSE indicates higher accuracy.
We provide two RMSE evaluations for each variable in the scaled (i.e., normalized) and original values.
The characteristic columns are calculated as scaled values. 

It is worth mentioning that the hydrology dataset contains a large number of meaningful zero values. For example, the variable $D$, named Snowpack, only has non-zero values in winter, whose water body will present as Soil Water (variable E) in the other seasons. However, the zeros do have meanings since they represent the variables vanishing.
Therefore, we perform double reconstructions simultaneously in the autoencoder: one for the variable's continuous values, and the other to be the non-zero indicator, evaluated by BCE and named Mask in the table. The RMSE performances in Table \ref{tab:tower} are obtained by multiplying these two results.
The percentage of non-zero values is also provided for each node.
These shallow RMSE values indicate the success of the reconstruction processes, which are in the range of $[0.01, 0.09]$, except node $F$, the Aquifer variable.
It has been known that the aquifer system modeling is still premature in the present hydrology area. So this is reasonable to infer that in the synthetic data, Aquifer is closer to random noise than other variables.

\begin{table*}[t]
\caption{Performance of Autoencoder Reconstruction.}
\label{tab:tower}
\vspace{-3mm}
\resizebox{1.9\columnwidth}{!}{%
\begin{tabular}{|
>{\columncolor[HTML]{EFEFEF}}c |c|l|l|l|l|l|l|l|l|}
\hline
\cellcolor[HTML]{C0C0C0}Variable & \cellcolor[HTML]{C0C0C0} Length & \multicolumn{1}{c|}{\cellcolor[HTML]{C0C0C0}Mean} & \multicolumn{1}{c|}{\cellcolor[HTML]{C0C0C0}Std} & \multicolumn{1}{c|}{\cellcolor[HTML]{C0C0C0}Min} & \multicolumn{1}{c|}{\cellcolor[HTML]{C0C0C0}Max} & \multicolumn{1}{c|}{\cellcolor[HTML]{C0C0C0}Non-Zero Rate\%} & \multicolumn{1}{c|}{\cellcolor[HTML]{C0C0C0}RMSE on Scaled} & \multicolumn{1}{c|}{\cellcolor[HTML]{C0C0C0}RMSE on Original} & \multicolumn{1}{c|}{\cellcolor[HTML]{C0C0C0}BCE of Mask} \\ \hline
A & 5 & 1.8513 & 1.5496 & -3.3557 & 7.6809 & 87.54 & 0.093 & 0.871 & 0.095 \\ \hline
B & 4 & 0.7687 & 1.1353 & -3.3557 & 5.9710 & 64.52 & 0.076 & 0.678 & 1.132 \\ \hline
C & 2 & 1.0342 & 1.0025 & 0.0 & 6.2145 & 94.42 & 0.037 & 0.089 & 0.428 \\ \hline
D & 3 & 0.0458 & 0.2005 & 0.0 & 5.2434 & 11.40 & 0.015 & 0.679 & 0.445 \\ \hline
E & 2 & 3.1449 & 1.0000 & 0.0285 & 5.0916 & 100 & 0.058 & 3.343 & 0.643 \\ \hline
F & 4 & 0.3922 & 0.8962 & 0.0 & 8.6122 & 59.08 & 0.326 & 7.178 & 2.045 \\ \hline
G & 4 & 0.7180 & 1.1064 & 0.0 & 8.2551 & 47.87 & 0.045 & 0.81 & 1.327 \\ \hline
H & 4 & 0.7344 & 1.0193 & 0.0 & 7.6350 & 49.93 & 0.045 & 0.009 & 1.345 \\ \hline
I & 3 & 0.1432 & 0.6137 & 0.0 & 8.3880 & 21.66 & 0.035 & 0.009 & 1.672 \\ \hline
J & 1 & 0.0410 & 0.2000 & 0.0 & 7.8903 & 21.75 & 0.007 & 0.098 & 1.088 \\ \hline
\end{tabular}}
\end{table*}

\begin{table*}[t]
\caption{Brief Summary of the Latent Causal Discovery Results. }
\label{tab:discv}
\vspace{-3mm}
\begin{tabular}{|c|l|l|l|l|l|l|l|l|l|l|l|l|l|l|l|l|}
\hline
\cellcolor[HTML]{C0C0C0}Edge & \cellcolor[HTML]{FFCCC9}A$\Rightarrow$C & \cellcolor[HTML]{FFCCC9}B$\Rightarrow$D & \cellcolor[HTML]{FFCCC9}C$\Rightarrow$D & \cellcolor[HTML]{FFCCC9}C$\Rightarrow$G & \cellcolor[HTML]{FFCCC9}D$\Rightarrow$G & \cellcolor[HTML]{FFCCC9}G$\Rightarrow$J & \cellcolor[HTML]{FFCCC9}D$\Rightarrow$H & \cellcolor[HTML]{FFCCC9}H$\Rightarrow$J & \cellcolor[HTML]{FFCE93}B$\Rightarrow$E & \cellcolor[HTML]{FFCE93}E$\Rightarrow$G & \cellcolor[HTML]{FFCE93}E$\Rightarrow$H & \cellcolor[HTML]{FFCE93}C$\Rightarrow$E & \cellcolor[HTML]{ABE9E7}E$\Rightarrow$F & \cellcolor[HTML]{ABE9E7}F$\Rightarrow$I & \cellcolor[HTML]{ABE9E7}I$\Rightarrow$J & \cellcolor[HTML]{ABE9E7}D$\Rightarrow$I \\ \hline
\cellcolor[HTML]{C0C0C0}KLD & 7.63 & 8.51 & 10.14 & 11.60 & 27.87 & 5.29 & 25.19 & 15.93 & 37.07 & 39.13 & 39.88 & 46.58 & 53.68 & 45.64 & 17.41 & 75.57 \\ \hline
\rowcolor[HTML]{FFFFFF} 
\cellcolor[HTML]{C0C0C0}Gain & 7.63 & 8.51 & 1.135 & 11.60 & 2.454 & 5.29 & 25.19 & 0.209 & 37.07 & -5.91 & -3.29 & 2.677 & 53.68 & 45.64 & 0.028 & 3.384 \\ \hline
\end{tabular}
\vspace{-1mm}
\end{table*}

\subsection{Latent Causal Effects Learning Test}
%\vspace{-1.5mm}
The latent causal effect learning is evaluated on pair-wise causations.
And by stacking them, we successfully recover the ground truth graphical causal structure displayed in Fig~\ref{fig:stream}.
To compare the changing of the  reconstructing performance by each stacking, we construct the Table \ref{tab:unit}
for each variable that has ever been the \emph{result} node of pair-wise causations. 
For example, node $G$ has three possible causes $C$, $D$, and $E$, so we can build four causal effect models in total: the individual pair-wise causations
$C\Rightarrow G$, $D\Rightarrow G$, $E\Rightarrow G$, and the stacked causal effects $CDE\Rightarrow G$. 
For convenience, we call them ``\emph{pair-effect}'' and ``\emph{stacking-effect}'' respectively in the following.
In Table \ref{tab:unit} we also list the initial reconstruction performance for each variable, as the comparison baseline, in the column named ``Variable Reconstruction (initial performance)''.

%\vspace{-2mm}
\begin{figure*}[h!]
\centering
\includegraphics[scale=0.65]{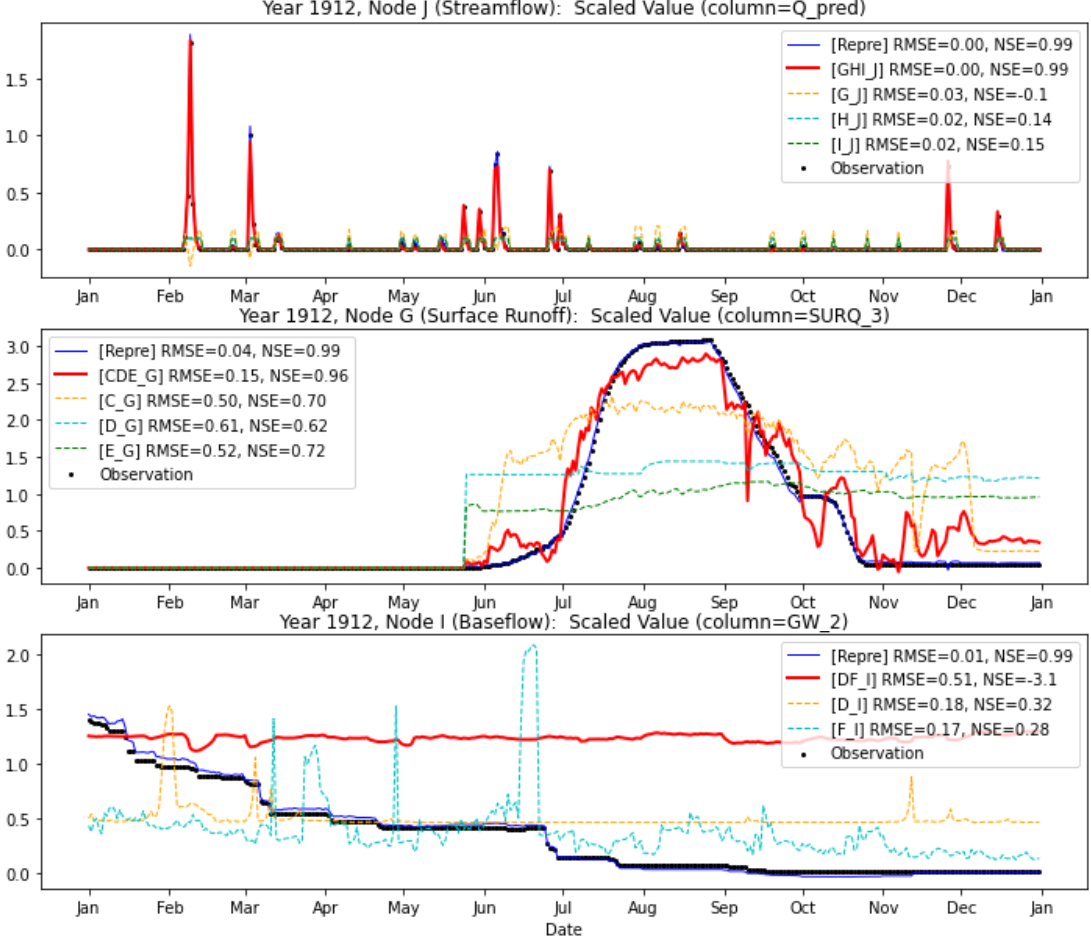}
\vspace*{-3mm}
\caption{Performance of Latent Causal Effect Reconstruction Examples. (evaluated as data simulation accuracy)}
\label{fig:G}
\vspace{-2mm}
\end{figure*}

% %\vspace{-2mm}
% \begin{figure*}[h!]
% \centering
% \includegraphics[scale=0.74]{fig_node_j.png}
% \vspace*{-4mm}
% \caption{ Data simulations of node J. }
% \label{fig:J}
% %\vspace{-5mm}
% \end{figure*}

There are three different optimization tasks in the latent causal effect learning process (refer to Section~\ref{sec:method_3}), among which both the \emph{second} and \emph{third} one are about the \emph{result} node. Therefore, in Table \ref{tab:unit}, we give out the performances of these two roles for each node: the role in the \emph{third} optimization is shown as the columns named ``Variable Reconstruction (as result node)''; and the role of the \emph{second} one is in the columns named ``Latent Causal Effect Reconstruction'', which represent the causal effect learning performances of RNN models, along with the KLD metrics to display the learned causality strength (the lower is stronger).

The KLD metric differences among variables are easy to be observed. 
Variable $J$ has the smallest KLD values, which means that $G$ (Surface Runoff), $H$ (Lateral), and $I$ (Baseflow) significantly cause $J$ (Streamflow).
While the weaker causal relations present as higher values of KLD. For example, variable $I$ can hardly be predictable given $D$ and $F$.
For \emph{result} nodes, $D$, $E$, $J$, the \emph{stacking-effect} causal strength is in the intermediate range of \emph{pair-effect} strength, which implies that the internal associative relationship inside their cause nodes is ambiguous.
For $G$ and $H$, the \emph{stacking-effect} causal strength KLD is lower than all the \emph{pair-effect} ones. It is reasonable to infer that the \emph{stacking-effect} model has captured additional causal information from the associative relationships among their causes nodes.
Besides, the KLD metric also tells which cause node contributes the most to the latent causal effect. For example, $C\Rightarrow G$ strength is closer to $CDE\Rightarrow G$ than the other causes nodes, which indicates $C$ to be the strongest source in this causal effect. 

To better present the reconstruction performances of the stacked causal structure, we illustrate their results as data simulations in Figure~\ref{fig:G}, which includes three different nodes, $J$, $G$, and $I$ in the same synthetic year. For each node, the initial variable reconstruction performance is plotted as the blue line, and all the latent causal effects we have learned are displayed in different colors.
Besides RMSE, here we also use another metric, NSE (Nash–Sutcliffe model efficiency coefficient) to be the hydrology-meaningful accuracy evaluation, where NSE $=1$ defines the best prediction.
The three initial variable reconstructions (blue lines) in Figure \ref{fig:G} almost reach the best performance, as they overlapped with the ground truth observations (black dots).
The red lines indicate simulations from \emph{stacking-effect}, whose performances are highly consistent with our analysis above: Node $J$ has the best prediction; node $I$ can hardly be predictable; and node G is predicted best by $CDE$ causes combination, among which $C$ provides the strongest causality.

%\vspace{-6mm}
\subsection{Causal Structure Discovery Test}

The latent causal discovery is based on the KLD causality strength evaluation.
\ref{tab:discv} displays the edges' discovered order and their corresponding KLD and KLD-Gains. Here the table cells are colored according to the different tiers of causality, consistent with the ground truth shown in Figure \ref{fig:stream}.
Since the discovery algorithm successfully distinguished the 3 tiers, the proposed KLD metric has been proven to work effectively.

Due to limited space, the results in Table \ref{tab:discv} is only a brief version, whose complete version is provided in Appendix A Table 1, which includes all results during the causal discovery, organized by the round of detection.
As a baseline comparison, we also performed the conventional FGES (fast greedy equivalence search) causal discovery method in $10$ cross-validation. Their results are given in Appendix A Table 2, with apparently worse performance than the proposed CRL method.

\vspace{-1mm}
\section{Conclusion}\label{sec:conclusion}
Recently, the application ChatGPT has become very popular, and we are interested in how well it can replace humans in doing intellectual work.
On the contrary, we barely hear AI can replace doctors in making prescription decisions or cracking the genetic code automatically, although they have much fewer computational complexities than learning tens of languages, computing pixels of movies, or evaluating billions of possibilities in the Go game.
So, what sealed AI's capability when learning structural data?

Understanding the timeline is a fundamental difference between humans and AI, distinguishing causality from correlation. We must fully consider the individual-level heterogeneity (i.e., the personalized time-passing speed differences) before feeding the structural data to AI to establish generalizable causal models. We hope the proposed CRL can lead to the next trend.

% In this paper, we initially raised the geometric meaning of causal graphs, and accordingly, analyzed the existence of Causal Representation Bias (CRB), which has been rarely noticed in causality machine learning.
% Such a blind spot has hindered machine-learning development in practical applications.

% Specifically, deep learning can achieve great performance, but the black-box nature makes it easier to overlook potential CRBs, and thus is restricted by model generalizability.
% On the other hand, causal inference analytics focuses on the intra-timeline relationship but lacks the geometric global view to realize the inter-timeline defines the individual-level features.

% In this paper, we proposed a generic learning framework CRL, demonstrated its realization, and experimentally proved its feasibility.
% We suggest that if the causal relationship potentially contains multiple relatively independent timelines or the cohort data is a mixture of populations, using CRL can help to avoid potential CRBs.

\begin{table*}[t]
\caption{The latent causal effect learning performances.}
\label{tab:unit}
\vspace{-3mm}
\resizebox{1.99\columnwidth}{!}{%
\begin{tabular}{|l|lll|l|lll|llll|}
\hline
\rowcolor[HTML]{C0C0C0} 
\cellcolor[HTML]{C0C0C0}{\color[HTML]{000000} } & \multicolumn{3}{l|}{\cellcolor[HTML]{C0C0C0}{\color[HTML]{000000} \begin{tabular}[c]{@{}l@{}}\ \ \ \ \ \ Variable Reconstruction\\ \ \ \ \ \ \ \ (initial performance)\end{tabular}}} & \cellcolor[HTML]{C0C0C0}{\color[HTML]{000000} } & \multicolumn{3}{l|}{\cellcolor[HTML]{C0C0C0}{\color[HTML]{000000}  \begin{tabular}[c]{@{}l@{}}\ \ \ \ \ \ Variable Reconstruction\\ \ \ \ \ \ \ \ (as result node)\end{tabular}}} & \multicolumn{4}{l|}{\cellcolor[HTML]{C0C0C0}{\color[HTML]{000000} \ \ \ \ \ \ \ \ \ Latent Causal Effect Reconstruction}} \\ \cline{2-4} \cline{6-12} 
\rowcolor[HTML]{C0C0C0} 
\cellcolor[HTML]{C0C0C0}{\color[HTML]{000000} } & \multicolumn{2}{l|}{\cellcolor[HTML]{C0C0C0}{\color[HTML]{000000}\ \ \ \ \ \  \ \ \ \ RMSE}} & {\color[HTML]{000000} BCE} & \cellcolor[HTML]{C0C0C0}{\color[HTML]{000000} } & \multicolumn{2}{l|}{\cellcolor[HTML]{C0C0C0}\ \ \ \ \ \ \ \ \ \ RMSE} & BCE & \multicolumn{2}{l|}{\cellcolor[HTML]{C0C0C0} \ \ \ \ \ \ \ \ \ RMSE} & \multicolumn{1}{l|}{\cellcolor[HTML]{C0C0C0}BCE} & KLD \\ \cline{2-4} \cline{6-12} 
\rowcolor[HTML]{C0C0C0} 
\multirow{-3}{*}{\cellcolor[HTML]{C0C0C0}{\color[HTML]{000000} \begin{tabular}[c]{@{}l@{}}Variable\\ \ \end{tabular}}} & \multicolumn{1}{l|}{\cellcolor[HTML]{C0C0C0}\begin{tabular}[c]{@{}l@{}}on Scaled\\ \ \ Values\end{tabular}} & \multicolumn{1}{l|}{\cellcolor[HTML]{C0C0C0}\begin{tabular}[c]{@{}l@{}}on Original\\ \ \ Values\end{tabular}} & Mask & \multirow{-3}{*}{\cellcolor[HTML]{C0C0C0}{\color[HTML]{000000} \begin{tabular}[c]{@{}l@{}}Cause\\ Node\end{tabular}}} & \multicolumn{1}{l|}{\cellcolor[HTML]{C0C0C0}\begin{tabular}[c]{@{}l@{}}on Scaled\\ \ \ Values\end{tabular}} & \multicolumn{1}{l|}{\cellcolor[HTML]{C0C0C0}\begin{tabular}[c]{@{}l@{}}on Original\\ \ \ Values\end{tabular}} & Mask & \multicolumn{1}{l|}{\cellcolor[HTML]{C0C0C0}\begin{tabular}[c]{@{}l@{}}on Scaled\\ \ \ Values\end{tabular}} & \multicolumn{1}{l|}{\cellcolor[HTML]{C0C0C0}\begin{tabular}[c]{@{}l@{}}on Original\\ \ \ Values\end{tabular}} & \multicolumn{1}{l|}{\cellcolor[HTML]{C0C0C0}Mask} & \begin{tabular}[c]{@{}l@{}}(in latent\\ \ \ space)\end{tabular} \\ \hline
\cellcolor[HTML]{EFEFEF}C & \multicolumn{1}{l|}{\cellcolor[HTML]{EFEFEF}0.037} & \multicolumn{1}{l|}{\cellcolor[HTML]{EFEFEF}0.089} & \cellcolor[HTML]{EFEFEF}0.428 & A & \multicolumn{1}{l|}{0.0295} & \multicolumn{1}{l|}{0.0616} & 0.4278 & \multicolumn{1}{l|}{0.1747} & \multicolumn{1}{l|}{0.3334} & \multicolumn{1}{l|}{0.4278} & 7.6353 \\ \hline
\cellcolor[HTML]{EFEFEF} & \multicolumn{1}{l|}{\cellcolor[HTML]{EFEFEF}} & \multicolumn{1}{l|}{\cellcolor[HTML]{EFEFEF}} & \cellcolor[HTML]{EFEFEF} & BC & \multicolumn{1}{l|}{0.0350} & \multicolumn{1}{l|}{1.0179} & 0.1355 & \multicolumn{1}{l|}{0.0509} & \multicolumn{1}{l|}{1.7059} & \multicolumn{1}{l|}{0.1285} & 9.6502 \\ \cline{5-5}
\cellcolor[HTML]{EFEFEF} & \multicolumn{1}{l|}{\cellcolor[HTML]{EFEFEF}} & \multicolumn{1}{l|}{\cellcolor[HTML]{EFEFEF}} & \cellcolor[HTML]{EFEFEF} & B & \multicolumn{1}{l|}{0.0341} & \multicolumn{1}{l|}{1.0361} & 0.1693 & \multicolumn{1}{l|}{0.0516} & \multicolumn{1}{l|}{1.7737} & \multicolumn{1}{l|}{0.1925} & 8.5147 \\ \cline{5-5}
\multirow{-3}{*}{\cellcolor[HTML]{EFEFEF}D} & \multicolumn{1}{l|}{\multirow{-3}{*}{\cellcolor[HTML]{EFEFEF}0.015}} & \multicolumn{1}{l|}{\multirow{-3}{*}{\cellcolor[HTML]{EFEFEF}0.679}} & \multirow{-3}{*}{\cellcolor[HTML]{EFEFEF}0.445} & C & \multicolumn{1}{l|}{0.0331} & \multicolumn{1}{l|}{0.9818} & 0.3404 & \multicolumn{1}{l|}{0.0512} & \multicolumn{1}{l|}{1.7265} & \multicolumn{1}{l|}{0.3667} & 10.149 \\ \hline
\cellcolor[HTML]{EFEFEF} & \multicolumn{1}{l|}{\cellcolor[HTML]{EFEFEF}} & \multicolumn{1}{l|}{\cellcolor[HTML]{EFEFEF}} & \cellcolor[HTML]{EFEFEF} & BC & \multicolumn{1}{l|}{0.4612} & \multicolumn{1}{l|}{26.605} & 0.6427 & \multicolumn{1}{l|}{0.7827} & \multicolumn{1}{l|}{45.149} & \multicolumn{1}{l|}{0.6427} & 39.750 \\ \cline{5-5}
\cellcolor[HTML]{EFEFEF} & \multicolumn{1}{l|}{\cellcolor[HTML]{EFEFEF}} & \multicolumn{1}{l|}{\cellcolor[HTML]{EFEFEF}} & \cellcolor[HTML]{EFEFEF} & B & \multicolumn{1}{l|}{0.6428} & \multicolumn{1}{l|}{37.076} & 0.6427 & \multicolumn{1}{l|}{0.8209} & \multicolumn{1}{l|}{47.353} & \multicolumn{1}{l|}{0.6427} & 37.072 \\ \cline{5-5}
\multirow{-3}{*}{\cellcolor[HTML]{EFEFEF}E} & \multicolumn{1}{l|}{\multirow{-3}{*}{\cellcolor[HTML]{EFEFEF}0.058}} & \multicolumn{1}{l|}{\multirow{-3}{*}{\cellcolor[HTML]{EFEFEF}3.343}} & \multirow{-3}{*}{\cellcolor[HTML]{EFEFEF}0.643} & C & \multicolumn{1}{l|}{0.5212} & \multicolumn{1}{l|}{30.065} & 1.2854 & \multicolumn{1}{l|}{0.7939} & \multicolumn{1}{l|}{45.791} & \multicolumn{1}{l|}{1.2854} & 46.587 \\ \hline
\cellcolor[HTML]{EFEFEF}F & \multicolumn{1}{l|}{\cellcolor[HTML]{EFEFEF}0.326} & \multicolumn{1}{l|}{\cellcolor[HTML]{EFEFEF}7.178} & \cellcolor[HTML]{EFEFEF}2.045 & E & \multicolumn{1}{l|}{0.4334} & \multicolumn{1}{l|}{8.3807} & 3.0895 & \multicolumn{1}{l|}{0.4509} & \multicolumn{1}{l|}{5.9553} & \multicolumn{1}{l|}{3.0895} & 53.680 \\ \hline
\cellcolor[HTML]{EFEFEF} & \multicolumn{1}{l|}{\cellcolor[HTML]{EFEFEF}} & \multicolumn{1}{l|}{\cellcolor[HTML]{EFEFEF}} & \cellcolor[HTML]{EFEFEF} & CDE & \multicolumn{1}{l|}{0.0538} & \multicolumn{1}{l|}{0.9598} & 0.0878 & \multicolumn{1}{l|}{0.1719} & \multicolumn{1}{l|}{3.5736} & \multicolumn{1}{l|}{0.1340} & 8.1360 \\ \cline{5-5}
\cellcolor[HTML]{EFEFEF} & \multicolumn{1}{l|}{\cellcolor[HTML]{EFEFEF}} & \multicolumn{1}{l|}{\cellcolor[HTML]{EFEFEF}} & \cellcolor[HTML]{EFEFEF} & C & \multicolumn{1}{l|}{0.1057} & \multicolumn{1}{l|}{1.4219} & 0.1078 & \multicolumn{1}{l|}{0.2996} & \multicolumn{1}{l|}{4.6278} & \multicolumn{1}{l|}{0.1362} & 11.601 \\ \cline{5-5}
\cellcolor[HTML]{EFEFEF} & \multicolumn{1}{l|}{\cellcolor[HTML]{EFEFEF}} & \multicolumn{1}{l|}{\cellcolor[HTML]{EFEFEF}} & \cellcolor[HTML]{EFEFEF} & D & \multicolumn{1}{l|}{0.1773} & \multicolumn{1}{l|}{3.6083} & 0.1842 & \multicolumn{1}{l|}{0.4112} & \multicolumn{1}{l|}{8.0841} & \multicolumn{1}{l|}{0.2228} & 27.879 \\ \cline{5-5}
\multirow{-4}{*}{\cellcolor[HTML]{EFEFEF}G} & \multicolumn{1}{l|}{\multirow{-4}{*}{\cellcolor[HTML]{EFEFEF}0.045}} & \multicolumn{1}{l|}{\multirow{-4}{*}{\cellcolor[HTML]{EFEFEF}0.81}} & \multirow{-4}{*}{\cellcolor[HTML]{EFEFEF}1.327} & E & \multicolumn{1}{l|}{0.1949} & \multicolumn{1}{l|}{4.7124} & 0.1482 & \multicolumn{1}{l|}{0.5564} & \multicolumn{1}{l|}{10.852} & \multicolumn{1}{l|}{0.1877} & 39.133 \\ \hline
\cellcolor[HTML]{EFEFEF} & \multicolumn{1}{l|}{\cellcolor[HTML]{EFEFEF}} & \multicolumn{1}{l|}{\cellcolor[HTML]{EFEFEF}} & \cellcolor[HTML]{EFEFEF} & DE & \multicolumn{1}{l|}{0.0889} & \multicolumn{1}{l|}{0.0099} & 2.5980 & \multicolumn{1}{l|}{0.3564} & \multicolumn{1}{l|}{0.0096} & \multicolumn{1}{l|}{2.5980} & 21.905 \\ \cline{5-5}
\cellcolor[HTML]{EFEFEF} & \multicolumn{1}{l|}{\cellcolor[HTML]{EFEFEF}} & \multicolumn{1}{l|}{\cellcolor[HTML]{EFEFEF}} & \cellcolor[HTML]{EFEFEF} & D & \multicolumn{1}{l|}{0.0878} & \multicolumn{1}{l|}{0.0104} & 0.0911 & \multicolumn{1}{l|}{0.4301} & \multicolumn{1}{l|}{0.0095} & \multicolumn{1}{l|}{0.0911} & 25.198 \\ \cline{5-5}
\multirow{-3}{*}{\cellcolor[HTML]{EFEFEF}H} & \multicolumn{1}{l|}{\multirow{-3}{*}{\cellcolor[HTML]{EFEFEF}0.045}} & \multicolumn{1}{l|}{\multirow{-3}{*}{\cellcolor[HTML]{EFEFEF}0.009}} & \multirow{-3}{*}{\cellcolor[HTML]{EFEFEF}1.345} & E & \multicolumn{1}{l|}{0.1162} & \multicolumn{1}{l|}{0.0105} & 0.1482 & \multicolumn{1}{l|}{0.5168} & \multicolumn{1}{l|}{0.0097} & \multicolumn{1}{l|}{3.8514} & 39.886 \\ \hline
\cellcolor[HTML]{EFEFEF} & \multicolumn{1}{l|}{\cellcolor[HTML]{EFEFEF}} & \multicolumn{1}{l|}{\cellcolor[HTML]{EFEFEF}} & \cellcolor[HTML]{EFEFEF} & DF & \multicolumn{1}{l|}{0.0600} & \multicolumn{1}{l|}{0.0103} & 3.4493 & \multicolumn{1}{l|}{0.1158} & \multicolumn{1}{l|}{0.0099} & \multicolumn{1}{l|}{3.4493} & 49.033 \\ \cline{5-5}
\cellcolor[HTML]{EFEFEF} & \multicolumn{1}{l|}{\cellcolor[HTML]{EFEFEF}} & \multicolumn{1}{l|}{\cellcolor[HTML]{EFEFEF}} & \cellcolor[HTML]{EFEFEF} & D & \multicolumn{1}{l|}{0.1212} & \multicolumn{1}{l|}{0.0108} & 3.0048 & \multicolumn{1}{l|}{0.2073} & \multicolumn{1}{l|}{0.0108} & \multicolumn{1}{l|}{3.0048} & 75.577 \\ \cline{5-5}
\multirow{-3}{*}{\cellcolor[HTML]{EFEFEF}I} & \multicolumn{1}{l|}{\multirow{-3}{*}{\cellcolor[HTML]{EFEFEF}0.035}} & \multicolumn{1}{l|}{\multirow{-3}{*}{\cellcolor[HTML]{EFEFEF}0.009}} & \multirow{-3}{*}{\cellcolor[HTML]{EFEFEF}1.672} & F & \multicolumn{1}{l|}{0.0540} & \multicolumn{1}{l|}{0.0102} & 3.4493 & \multicolumn{1}{l|}{0.0948} & \multicolumn{1}{l|}{0.0098} & \multicolumn{1}{l|}{3.4493} & 45.648 \\ \hline
\cellcolor[HTML]{EFEFEF} & \multicolumn{1}{l|}{\cellcolor[HTML]{EFEFEF}} & \multicolumn{1}{l|}{\cellcolor[HTML]{EFEFEF}} & \cellcolor[HTML]{EFEFEF} & GHI & \multicolumn{1}{l|}{0.0052} & \multicolumn{1}{l|}{0.0742} & 0.2593 & \multicolumn{1}{l|}{0.0090} & \multicolumn{1}{l|}{0.1269} & \multicolumn{1}{l|}{0.2937} & 5.5300 \\ \cline{5-5}
\cellcolor[HTML]{EFEFEF} & \multicolumn{1}{l|}{\cellcolor[HTML]{EFEFEF}} & \multicolumn{1}{l|}{\cellcolor[HTML]{EFEFEF}} & \cellcolor[HTML]{EFEFEF} & G & \multicolumn{1}{l|}{0.0077} & \multicolumn{1}{l|}{0.1085} & 0.4009 & \multicolumn{1}{l|}{0.0099} & \multicolumn{1}{l|}{0.1390} & \multicolumn{1}{l|}{0.4375} & 5.2924 \\ \cline{5-5}
\cellcolor[HTML]{EFEFEF} & \multicolumn{1}{l|}{\cellcolor[HTML]{EFEFEF}} & \multicolumn{1}{l|}{\cellcolor[HTML]{EFEFEF}} & \cellcolor[HTML]{EFEFEF} & H & \multicolumn{1}{l|}{0.0159} & \multicolumn{1}{l|}{0.2239} & 0.4584 & \multicolumn{1}{l|}{0.0393} & \multicolumn{1}{l|}{0.5520} & \multicolumn{1}{l|}{0.4938} & 15.930 \\ \cline{5-5}
\multirow{-4}{*}{\cellcolor[HTML]{EFEFEF}J} & \multicolumn{1}{l|}{\multirow{-4}{*}{\cellcolor[HTML]{EFEFEF}0.007}} & \multicolumn{1}{l|}{\multirow{-4}{*}{\cellcolor[HTML]{EFEFEF}0.098}} & \multirow{-4}{*}{\cellcolor[HTML]{EFEFEF}1.088} & I & \multicolumn{1}{l|}{0.0308} & \multicolumn{1}{l|}{0.4328} & 0.3818 & \multicolumn{1}{l|}{0.0397} & \multicolumn{1}{l|}{0.5564} & \multicolumn{1}{l|}{0.3954} & 17.410 \\ \hline
\end{tabular}
}
\vspace{-2mm}
\end{table*}

%\bibliographystyle{ACM-Reference-Format}
%\bibliography{acmart}

%\newpage

\begin{appendices}
\section{Complete Experimental Results of Causal Discovery}

\end{appendices}

\pagebreak


\section*{Supplementary material}

| # | | | | | | | | | | | | | |
|---|---|---|---|---|---|---|---|---|---|---|---|---|---|
| #1 | A→C 7.6354 | A→D 19.7407 | A→E 60.1876 | A→F 119.7730 | **B→C 8.4753** | B→D 8.5147 | B→E 65.9335 | B→F 132.7717 | | | | | |
| #2 | A→D 19.7407 | A→E 60.1876 | A→F 119.7730 | B→D 8.5147 | B→E 65.9335 | B→F 132.7717 | C→D 1.1355 | C→E 10.1490 | C→F 46.5876 | C→G 11.6012 | C→H 39.2361 | C→I 95.1564 | |
| #3 | **A→D 9.7357** | A→E 60.1876 | A→F 119.7730 | B→E 65.9335 | B→F 132.7717 | C→D 1.1355 | C→E 46.5876 | C→F 111.2978 | C→G 11.6012 | C→H 39.2361 | C→I 95.1564 | D→E 63.7348 | D→F 123.3203 | D→G 27.8798 | D→H 25.1988 | D→I 75.5775 |
| #4 | A→E 60.1876 | A→F 119.7730 | B→E 65.9335 | B→F 132.7717 | C→E 46.5876 | C→F 111.2978 | C→G **11.6012** | C→H 39.2361 | C→I 95.1564 | D→E 63.7348 | D→F 123.3203 | D→G 27.8798 | D→H 25.1988 | D→I 75.5775 |
| #5 | A→E 60.1876 | A→F 119.7730 | B→E 65.9335 | B→F 132.7717 | C→E 46.5876 | C→F 111.2978 | C→H **11.6012** | C→I 95.1564 | D→E 63.7348 | D→F 123.3203 | D→G 27.8798 | D→H 25.1988 | D→I 75.5775 | G→J 5.2924 |
| #6 | A→E 60.1876 | A→F 119.7730 | B→E 65.9335 | B→F 132.7717 | C→E 46.5876 | C→F 111.2978 | C→H 39.2361 | C→I 95.1564 | D→E 63.7348 | D→F 123.3203 | D→G **2.4540** | D→H 25.1988 | D→I 75.5775 | G→J 5.2924 |
| #7 | A→E 60.1876 | A→F 119.7730 | B→E 65.9335 | B→F 132.7717 | C→E 46.5876 | C→F 111.2978 | C→H **39.2361** | C→I 95.1564 | D→E 63.7348 | D→F 123.3203 | D→H 25.1988 | D→I 75.5775 |
| #8 | A→E 60.1876 | A→F 119.7730 | B→E 65.9335 | B→F 132.7717 | C→E 46.5876 | C→F 111.2978 | C→I 95.1564 | D→E 63.7348 | D→F 123.3203 | D→I 75.5775 | H→J 0.2092 |
| #9 | **A→E 60.1876** | A→F 119.7730 | B→E 65.9335 | B→F 132.7717 | **C→E 46.5876** | C→I 95.1564 | D→E 63.7348 | D→F 123.3203 | D→I 75.5775 |
| #10 | A→F 119.7730 | **B→E −6.8372** | C→F 132.7717 | C→I 95.1564 | **D→E 17.0407** | D→F 123.3203 | D→I 75.5775 | E→G 53.6806 | E→H −5.9191 | E→I −3.2931 |
| #11 | A→F 119.7730 | B→F 132.7717 | C→F 111.2978 | D→F 123.3203 | D→I 75.5775 | **E→G** 53.6806 | **E→H −3.2931** | E→I 110.2558 |
| #12 | A→F 119.7730 | B→F 132.7717 | C→F 111.2978 | C→I 95.1564 | D→F 123.3203 | D→I 75.5775 | **E→F** 53.6806 | E→I 110.2558 |
| #13 | A→F 119.7730 | **B→F 132.7717** | C→F 111.2978 | C→I 95.1564 | D→F 123.3203 | D→I 75.5775 | **E→F 53.6806** | E→I 110.2558 |
| #14 | **A→F 119.7730** | **E→I 111.2978** | **F→I 45.6490** |
| #15 | C→I 95.1564 | D→I 75.5775 | **E→I 110.2558** | **I→J 0.0284** |
| #16 | **C→I 15.0222** | D→I 3.3845 |
| #16 | **C→I 15.0222** | **D→I 3.3845** |

Table 1: The Complete Results of Heuristic Causal Discovery in latent space. Each row stands for a round of detection, with "#" identifying the round number, and all candidate edges are listed with their KLD gains as below. 1) Green cells: the newly detected edges. 2) Red cells: the selected edge. 3) Blue cells: the trimmed edges accordingly.

1

Table 2: Average performance of 10-Fold FGES (Fast Greedy Equivalence Search) causal discovery, with the prior knowledge that each node can only cause the other nodes with the same or greater depth with it. An edge means connecting two attributes from two different nodes, respectively. Thus, the number of possible edges between two nodes is the multiplication of the numbers of their attributes, i.e., the lengths of their data vectors.
(All experiments are performed with 6 different Independent-Test kernels, including chi-square-test, prob-test, disc-bic-test, fisher-z-test, mvplr-test, d-sep-test. But their results turn out to be identical.)

| Cause Node | A | B | | C | | | D | | | E | | | F | G | H | I |
|---|---|---|---|---|---|---|---|---|---|---|---|---|---|---|---|---|
| True Causation | A→C | B→D | B→E | C→D | C→E | C→G | D→G | D→H | D→I | E→G | E→H | E→F | F→I | G→J | H→J | I→J |
| Number of Edges | 16 | 24 | 16 | 6 | 4 | 8 | 12 | 12 | 9 | 8 | 8 | 8 | 12 | 4 | 4 | 3 |
| Probability of Missing | 0.038889 | 0.125 | 0.125 | 0.062 | 0.06875 | 0.039286 | 0.069048 | 0.2 | 0.142857 | 0.003571 | 0.2 | 0.3 | 0.142857 | 0.0 | 0.072727 | 0.030303 |
| Wrong Causation | | | | C→F | | | D→E | D→F | | | | | F→G | G→H | G→I | H→I |
| Times of Wrongly Discovered | | | | 5.6 | | | 1.2 | 0.8 | | | | | 5.0 | 8.2 | 3.0 | 2.8 |

Table 3: Brief Results of the Heuristic Causal Discovery in latent space, identical with Table 3 in the paper body, for better comparison to the traditional FGES methods results on this page.
The edges are arranged in detected order (from left to right) and their measured causal strengths in each step are shown below correspondingly.
Causal strength is measured by KLD values (less is stronger). Each round of detection is pursuing the least KLD gain globally. All evaluations are in 4-Fold validation average values. Different colors represent the ground truth causality strength tiers (referred to the Figure 10 in the paper body).

| Causation | A→C | B→D | C→D | C→G | D→G | G→J | D→H | H→J | C→E | B→E | E→G | E→H | E→F | F→I | I→J | D→I |
|---|---|---|---|---|---|---|---|---|---|---|---|---|---|---|---|---|
| KLD | 7.63 | 8.51 | 10.14 | 11.60 | 27.87 | 5.29 | 25.19 | 15.93 | 46.58 | 65.93 | 39.13 | 39.88 | 53.68 | 45.64 | 17.41 | 75.57 |
| Gain | 7.63 | 8.51 | 1.135 | 11.60 | 2.454 | 5.29 | 25.19 | 0.209 | 46.58 | -6.84 | -5.91 | -3.29 | 53.68 | 45.64 | 0.028 | 3.384 |



\end{document}